\documentclass[10pt,twocolumn,letterpaper]{article}

\usepackage{iccv}
\usepackage{times}
\usepackage{epsfig}
\usepackage{graphicx}
\usepackage{amsmath}
\usepackage{amssymb}
\usepackage{color,soul}
\usepackage{multirow}
\usepackage{enumitem}
\usepackage[normalem]{ulem}
\usepackage{algorithm}
\usepackage{xcolor}
\usepackage{listings}
\usepackage[font=small]{caption}
\usepackage[font=footnotesize]{subcaption}
\usepackage{stmaryrd}
\usepackage{colortbl}
\usepackage{booktabs}   

\definecolor{mygray}{gray}{.9}

\newcommand{\venue}[1]{{$_{\text{#1}}$}}

\usepackage[pagebackref=true,breaklinks=true,letterpaper=true,colorlinks,bookmarks=false]{hyperref}

\iccvfinalcopy 

\def\iccvPaperID{2044} 

\ificcvfinal\fi

\begin{document}

\title{Orthogonal Projection Loss}

\author{
    Kanchana Ranasinghe$^{\dagger}$, Muzammal Naseer$^{*}$,  Munawar Hayat$^ \mathsection$, Salman Khan$^{\dagger *}$,
    Fahad Shahbaz Khan$^{\dagger\ddagger}$ \\
   $^\dagger$Mohamed bin Zayed University of AI, UAE 
   \;\; $^*$Australian National University, Australia \\
    $^ \mathsection$Monash University, Australia \;\; $^\ddagger$Linköping University, Sweden\\
  \texttt{\small kanchana.ranasinghe@mbzuai.ac.ae} 
  }

\maketitle
\ificcvfinal\thispagestyle{empty}\fi

\begin{abstract}
\vspace{-0.5em}
Deep neural networks have achieved remarkable performance on a range of classification tasks, with softmax cross-entropy (CE) loss emerging as the de-facto objective function. The CE loss encourages features of a class to have a higher projection score on the true class-vector compared to the negative classes. However, this is a relative constraint and does not explicitly force different class features to be well-separated. Motivated by the observation that ground-truth class representations in CE loss are orthogonal (one-hot encoded vectors), we develop a novel loss function termed `Orthogonal Projection Loss' (OPL) which imposes orthogonality in the feature space. OPL augments the properties of CE loss and directly enforces inter-class separation alongside intra-class clustering in the feature space through orthogonality constraints on the mini-batch level. As compared to other alternatives of CE, OPL offers unique advantages \eg, no additional learnable parameters, does not require careful negative mining and is not sensitive to the batch size. Given the plug-and-play nature of OPL, we evaluate it on a diverse range of tasks including image recognition (CIFAR-100), large-scale classification (ImageNet), domain generalization (PACS) and few-shot learning (miniImageNet, CIFAR-FS, tiered-ImageNet and Meta-dataset) and demonstrate its effectiveness across the board. Furthermore, OPL offers better robustness against practical nuisances such as adversarial attacks and label noise. Code is available at:  {\small\url{https://github.com/kahnchana/opl}}.
\vspace{-1em}
\end{abstract}

\vspace{-1em}
\section{Introduction}

\global\csname @topnum\endcsname 0 
\begin{figure}[t]
	\begin{center}
		\includegraphics[width=0.9\linewidth]{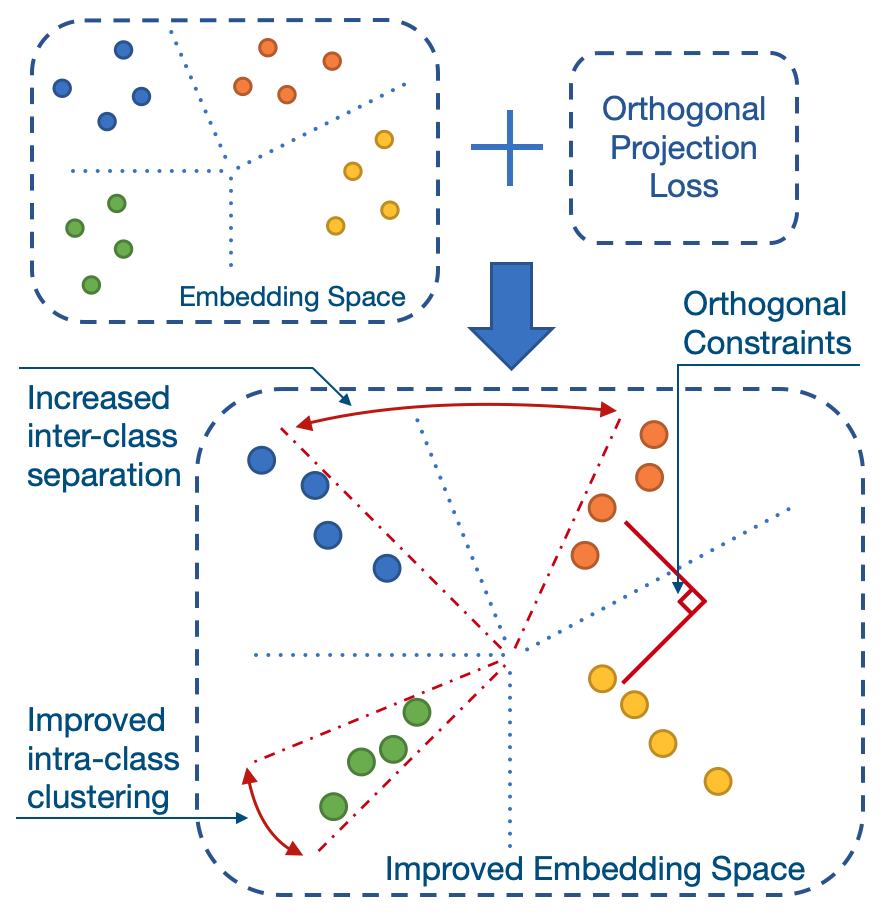}
	\end{center}
	\vspace{-4mm}
	\caption{\textbf{Orthogonal Projection Loss:} 
	During training of a deep neural network, within each mini-batch, OPL enforces separation between features of different class samples while clustering together features of the same class samples. OPL integrates well with softmax CE loss as it simply complements its intrinsic angular property, leading to consistent performance improvements on various classification tasks with a variety of DNN backbones.}
	\label{fig:opl_explanation}
\vspace{-5mm}
\end{figure}

Recent years have witnessed great success across a range of computer vision tasks owing to progress in deep neural networks (DNNs) \cite{lecun2015deep}. Effective loss functions for DNNs training have been a crucial component of these advancements \cite{goodfellow2016deep}. In particular, the softmax cross entropy (CE) loss, commonly used for tackling classification problems, has been pivotal for stable and efficient training of DNNs. 

Multiple variants of CE have been explored to enhance discriminativity and generalizability of feature representations learned during training.
Contrastive \cite{contrastive_loss} and triplet \cite{triplet_loss} loss functions are a common class of methods that have gained popularity on tasks requiring more discriminative features. At the same time, methods like centre loss \cite{centre_loss} and contrastive centre loss \cite{contrastive_centre_loss} have attempted to explicitly enforce inter-class separation and intra-class clustering through Euclidean margins between class prototypes. Angular margin based losses \cite{large_margin_softmax, sphere_face, arc_Face, cos_face, norm_face} compose another class of objective functions that increase inter-class margins through altering the logits prior to the CE loss. 

While these methods have proven successful at promoting better inter-class separation and intra-class compactness, they do possess certain drawbacks. Contrastive and triplet loss functions \cite{contrastive_loss, triplet_loss} are dependent on carefully designed negative mining procedures, which are both time-consuming and performance-sensitive. Methods based on centre loss \cite{centre_loss, contrastive_centre_loss}, that work together with CE loss, promote margins in Euclidean space which is counter-intuitive to the intrinsic angular separation enforced through CE loss \cite{sphere_face}. Further, these methods introduce additional learnable parameters in the form of new class centres. Angular margin based loss functions \cite{sphere_face, large_margin_softmax} which are highly successful for face recognition tasks, make strong assumptions for face embeddings to lie on the hypersphere manifold, which does not hold universally for all computer vision tasks \cite{negative_cam}. Some loss designs are also specific to certain architecture classes \eg, \cite{negative_cam} can only work with DNNs which output Class Activation Maps \cite{zhou2015cnnlocalization}. 

In this work, we explore a novel direction of simultaneously enforcing inter-class separation and intra-class clustering through orthogonality constraints on feature representations learned in the penultimate layer (Fig.~\ref{fig:opl_explanation}). We propose Orthogonal Projection Loss (OPL), which can be applied on the feature space of any DNN as a plug-and-play module. We are motivated by how image classification inherently assumes independent output classes and how orthogonality constraints in feature space go hand in hand with the one-hot encoded (orthogonal) label space used with CE. Furthermore, orthogonality constraints provide a definitive geometric structure in comparison to arbitrarily increasing margins which are prone to change depending on the selected batch, thus reducing sensitivity to batch composition. Finally, simply maximizing the margins can cause negative correlation between classes and thereby unnecessarily focus on well-separated classes while we tend to ensure independence between different class features to successfully disentangle the class-specific characteristics. 
 
Compared with contrastive loss functions \cite{contrastive_loss,triplet_loss}, OPL operates directly on mini-batches, eliminating the requirement of complex negative sample mining procedures. By enforcing orthogonality through computing dot-products between feature vectors, OPL provides a natural augmentation to the intrinsic angular property of CE, as opposed to methods \cite{centre_loss,contrastive_centre_loss,hayat2019gaussian} that enforce an Euclidean margin in feature space. Furthermore, OPL introduces no additional learnable parameters unlike \cite{centre_loss,LGM_loss,contrastive_centre_loss}, operates independent of model architecture unlike \cite{negative_cam}, and in contrast to losses operating on the hypersphere manifold \cite{sphere_face,large_margin_softmax,arc_Face,cos_face}, performs well on a wide range of tasks. Our main contributions are:

\begin{itemize}[topsep=1pt,itemsep=0pt,partopsep=1ex,parsep=1ex,leftmargin=*]
    \item We propose a novel loss, OPL, that directly enforces inter-class separation and intra-class clustering via orthogonality constraints with no learnable parameters.
    \item Our orthogonality constraints are efficiently formulated compared to existing methods \cite{lezama2018ole,sun2017svdnet}, allowing mini-batch processing without the need to explicitly obtain singular values. This leads to a simple vectorized implementation of OPL directly integrating with CE.
    \item We extensively evaluate on a diverse range of image classification tasks highlighting the discriminative ability of OPL. Further, our results on few-shot learning (FSL) and domain generalization (DG) datasets establish the transferability and generalizability of features learned with OPL. Finally, we establish the improved robustness of learned features to adversarial attacks and label noise.
\end{itemize}

\section{Related Work}

\noindent\textbf{Loss Functions.}
Loss functions play a central role in all deep learning based computer vision tasks. Recent works include reformulating the generic softmax calculation to limit the variance and range of logits during training \cite{rbf_loss}, constraining the feature space to follow specific distributions \cite{LGM_loss}, and heuristically altering margins targeting specific tasks like few-shot learning \cite{negative_margin, aml_paper},  class imbalance \cite{khan2019striking,khan2017cost,hayat2019gaussian,focal_loss} and zero-shot learning \cite{rahman2018polarity}. OPL optimizes a different objective of inter-class separation and intra-class clustering through orthogonalization in the feature space. 

\noindent\textbf{Generalizable Representations.} 
Recent works explore the transferability of features learned via supervised training \cite{how_transferable_paper}, \eg FSL \cite{rfs_paper, meta_baseline_paper, unravelling_meta} and DG \cite{RSC_paper, dg_class_margins} tasks. Tian \etal \cite{rfs_paper} establish a strong FSL baseline using only standard (non-episodic) supervised pre-training. Adaptation of supervised pre-trained models to the episodic evaluation setting of FSL tasks is explored in \cite{feat_paper, meta_baseline_paper}. Goldblum \etal \cite{unravelling_meta} show the importance of margin-based regularization methods for FSL. Our work differs by building on orthogonality constraints to learn more transferable features, and is more compatible with CE as opposed to \cite{unravelling_meta}. Multiple DG methods also explore constraints on feature spaces \cite{RSC_paper, dg_class_margins} to boost cross-domain performance. In particular, \cite{dg_class_margins} explores inter-class separation and intra-class clustering through contrastive and triplet loss functions. OPL improves on these while eliminating the need for compute expensive and complex sample mining procedures. 

\noindent\textbf{Orthogonality.}
Orthogonality of kernels in DNNs is well explored with an aim to diversify the learned weight vectors \cite{deep_hyperspherical, orthogonal_cnn}. The idea of orthogonality is also used for disentangled representations such as in \cite{orthogonal_face_age} and to stabilize network training since orthogonalization ensures energy preservation \cite{desjardins2015natural,rodriguez2016regularizing}. Orthogonal weight initializations have also shown their promise towards improving learning behaviours \cite{mishkin2015all,xie2017all}.  However, all of these works operate in the parameter space. Remarkably, the previous formulations to achieve orthogonality in the feature space generally depend on computing singular value decomposition \cite{lezama2018ole,sun2017svdnet}, which can be numerically unstable, difficult to estimate for rectangular matrices, and undergoes an iterative process \cite{bansal2018can}. In contrast, our orthogonal constraints are enforced in a novel manner, realized via decomposition on the sample-to-sample relationships within a mini-batch, while simultaneously avoiding tedious pair/triplet computations.

\section{Proposed Method}

Maximizing inter-class separation while enhancing intra-class compactness is highly desirable for classification. While the commonly used cross entropy (CE) loss encourages logits of the same class to be closer together, it does not enforce any margin amongst different classes. There have been multiple efforts to integrate max-margin learning with CE. For example, Large-margin softmax \cite{large_margin_softmax} enforces inter-class separability directly on the
dot-product similarity while SphereFace \cite{sphere_face} and ArcFace
\cite{arc_Face} enforce multiplicative and additive angular margins on
the hypersphere manifold, respectively. Directly enforcing max-margin constraints to enhance discriminability in the angular domain is 
ill-posed and requires approximations \cite{large_margin_softmax}. Some works turn to the Euclidean space to enhance feature space discrimination. For example, centre loss \cite{centre_loss} clusters penultimate layer features using Euclidean distance. Similarly, Affinity Loss \cite{hayat2019gaussian} forms uniformly shaped equi-distant class-wise clusters based upon Gaussian distances in the Euclid space. Margin-maximizing objective functions in the Euclid space are not ideally suited to work along-side CE loss, since CE seeks to separate output logits in the angular domain. By enforcing orthogonality constraints, our proposed OPL loss maximally separates intermediate features in the angular domain, thus complementing cross-entropy loss which enhances angular discriminability in the output space. In the following discussion, we revisit CE loss in the context of max-margin learning, and argue why OPL loss is ideally suited to supplement CE.

\subsection{Revisiting Softmax Cross Entropy Loss}

Consider a deep neural network $\mathcal{H}$, which can be decomposed into $\mathcal{H} = \mathcal{H}_{\phi} \cdot \mathcal{H}_{\theta}$, where $\mathcal{H}_{\phi}$ is the feature extraction module and $\mathcal{H}_{\theta}$ is the classification module. Given an input-output pair $\{\mathbf{x}, y\}$, let $\mathbf{f}=\mathcal{H}_{\phi}(\mathbf{x}), \mathbf{f} \in \mathbb{R}^d$ be the intermediate features and $\hat{y} = \mathcal{H}_{\theta}(\mathbf{f}), \hat{y} \in \mathbb{R}^k$ be the output predictions. For brevity, let us define the classification module as a linear layer $\mathcal{H}_{\theta}=\mathbf{W}=[\mathbf{w}_1,\cdots\mathbf{w}_c]$ with no unit-biases, where $\mathbf{w}_i \in \mathbb{R}^d, i=1\cdots c$ are class-wise learnable projection vectors for $c$ classes.
The traditional CE loss can then be defined in terms of discrepancy between the predicted $\hat y$ and ground-truth label $y$, by projecting the features $\mathbf{f} \in \mathbb{R}^d$ onto the weight matrix $\mathbf{W} \in \mathbb{R}^{d \times c}$.

\begin{align}
	\mathcal{L}_{\texttt{CE}}(\hat y, y)  &= - \log \ \frac{\exp{(\mathbf{f}^T \mathbf{w}_y)}}{\sum\limits_{j} \exp{(\mathbf{f}^T \mathbf{w}_j)} } \\
	\propto \sum\limits_{j \ne y} & \exp{(\mathbf{f}^T \mathbf{w}_j - \mathbf{f}^T\mathbf{w}_y}) \notag\\
	\propto  \sum\limits_{j \ne y} & \exp{(\left \|  \mathbf{f} \right \|_2 \left \| \mathbf{w}_j \right \|_2 \cos (\theta_{j}) } - \left \| \mathbf{f} \right \|_2 \left \| \mathbf{w}_y \right \|_2 \cos (\theta_{y})) \notag
\end{align}

\begin{figure}
	\begin{center}
		\includegraphics[width=0.48\linewidth]{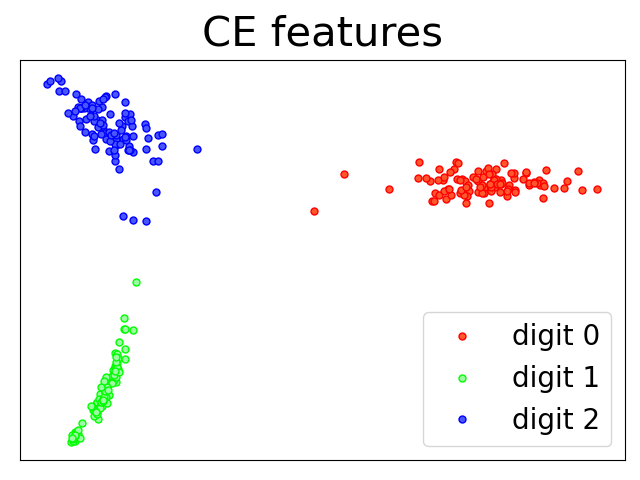}
		\includegraphics[width=0.48\linewidth]{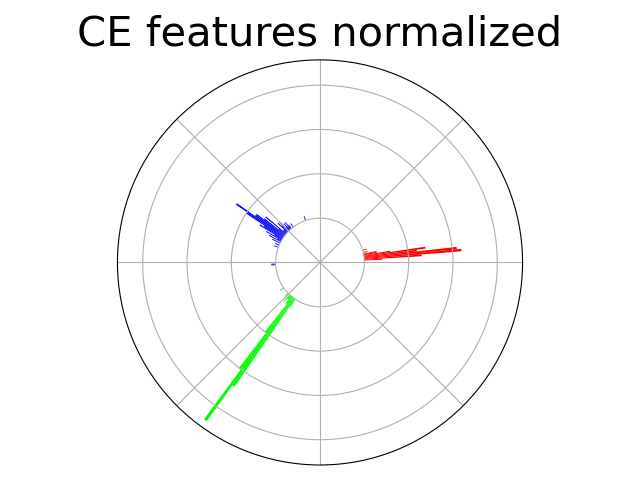}
		\includegraphics[width=0.48\linewidth]{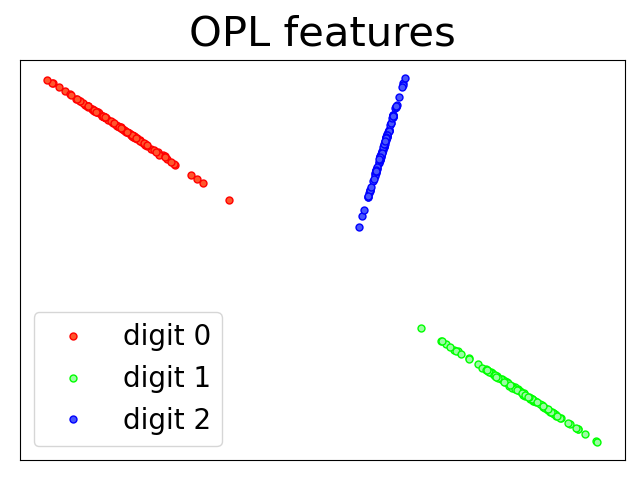}
		\includegraphics[width=0.48\linewidth]{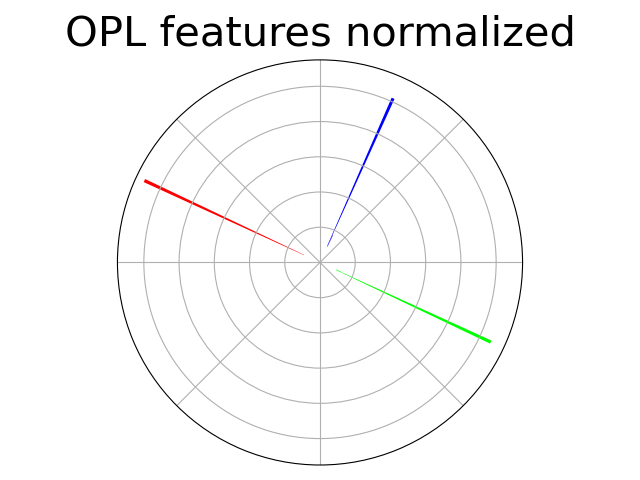}
	\end{center}
	\vspace{-4mm}
	\caption{\textbf{Feature space visualization for CE vs OPL:} Inter-class orthogonality enforced by OPL can be observed in this MNIST 2-D feature visualization. We plot only three classes to better illustrate in 2D the inter-class orthogonality achieved. Normalization refers to projection of vectors to a unit-hypersphere in feature-space, and the normalized plot contains a histogram for each angle.}
	\label{fig:ce_vs_opl_vis}
\vspace{-3mm}
\end{figure}

Since CE does not explicitly enforce any margin between each class pair, previous works have shown that the learned class regions for some class samples tend to be bigger compared to others \cite{sphere_face, arc_Face,hayat2019gaussian}. To counter this effect and ensure all classes are equally separated, efforts have been made to introduce a margin $m$ between different classes by modifying the term $\cos (\theta_y)$ as $\cos (m+\theta_y)$ for additive angular margin \cite{arc_Face}, $\cos (m\theta_y)$ for multiplicative angular margin \cite{sphere_face} and $m+\cos (\theta_y)$ as additive cosine margin \cite{cos_face}. The gradient propagation for such margin-based softmax loss formulations is hard, and previous works rely on approximations. Instead of introducing any margins to ensure uniform seperation between different classes, our proposed loss function simply enforces all classes to be orthogonal to each other with simultaneous clustering of within-class samples, using an efficient vectorized implementation with straightforward gradient computation and propagation. 

By considering the $\mathbf{w}_y \in \mathbf{W}$ vectors as individual class prototypes, the CE loss can be viewed as aligning the feature vectors $\mathbf{f}$ along its relevant class prototype. The cosine similarity in the form of the dot product ($\mathbf{f}^T \mathbf{w}_y$) gives CE an intrinsic angular property, which is observed in Fig.~\ref{fig:ce_vs_opl_vis} where features naturally separate in the polar coordinates with CE only. Moreover, during the standard SGD based optimization, the CE loss is applied on mini-batches. We note that there is no explicit enforcement of feature separation or clustering across multiple samples within the mini-batch. Given the opportunity to enforce such constraints since supervised training is commonly conducted adopting random mini-batch based iterations, we explore the possibility of within mini-batch constraints aimed at augmenting the intrinsic discriminative characteristics of CE loss.

\subsection{Orthogonal Projection Loss}
The CE loss with one-hot-encoded ground-truth vectors seeks to implicitly achieve orthogonality between different classes in the output space. Our proposed OPL loss, ameliorates CE loss, by enforcing class-wise orthogonality in the intermediate feature space. Given an input-output pair $\{\mathbf{x}_i, y_i\}$ in the dataset $\mathcal{D}$, let $\mathbf{f}_i = \mathcal{H}_{\phi}(\mathbf{x}_i)$ be the features output by an intermediate layer of the network. Our objective is to enforce constraints to cluster the features $\mathbf{f}_i \ \forall \mathbf{x}_i \in \mathcal{D}$ such that the features for different classes are orthogonal to each other and the features for the same class are similar. To this end, we define a unified loss function that simultaneously ensures intra-class clustering and inter-class orthogonality within a mini-batch as follows: 
\begin{align}
    s &= \sum_{\substack{i,j \in B \\ {y_i} = {y_j} }} \langle  \mathbf{f}_i, \mathbf{f}_j  \rangle \label{eq:loss_1} \\
    d &= \sum_{\substack{i,k \in B \\ {y_i} \ne {y_k} }} \langle  \mathbf{f}_i, \mathbf{f}_k  \rangle \label{eq:loss_2} \\
	\mathcal{L}_{\texttt{OPL}} &= (1-s) + |d|  \label{eq:loss_opl}
\end{align}
where $ \langle  \cdot \ , \cdot  \rangle$ is the cosine similarity operator applied on two vectors, $| \cdot |$ is the absolute value operator, and $B$ denotes mini-batch size. Note that the cosine similarity operator used in Eq. \ref{eq:loss_1} and \ref{eq:loss_2} involves normalization of features (projection to a unit hyper-sphere) as follows:
\begin{align}
    \langle  \mathbf{x}_i, \mathbf{x}_j \rangle = \frac{\mathbf{x}_i \cdot \mathbf{x}_j}{\|\mathbf{x}_i\|_2 \cdot \|\mathbf{x}_j\|_2} 
    \label{eq:cosine_sim}
\end{align}
where $|| \cdot ||_2$ refers to the $\ell_2$ norm operator. This normalization is key to aligning the outcome of OPL with the intrinsic angular property of CE loss. 

In Eq. \ref{eq:loss_opl}, our objective is to push $s$ towards 1 and $d$ towards 0. Since $1-s > 0$ already, we take the absolute value of $d$ given $d \in (-1, 1)$.  This in turn restricts the overall loss such that $\mathcal{L}_{\texttt{OPL}} \in (0, 3)$. When minimizing this overall loss, the first term $(1-s)$ will ensure clustering of same class samples, while the second term $|d|$ will ensure the orthogonality of different class samples. The loss can be implemented efficiently in a vectorized manner on the mini-batch level, avoiding any loops (see Algorithm \ref{alg:opl}). 

We further note that the ratio of contribution to the overall loss of each individual term in Eq. \ref{eq:loss_opl} can be controlled to re-prioritize between the two objectives of inter-class separation and intra-class compactness. While the unweighted combination of $s$ and $d$ alone performs well, specific use-cases could benefit from weighted combinations. We reformulate Eq. \ref{eq:loss_opl} as follows:
\begin{align}
    \mathcal{L}_{\texttt{OPL}} &= (1-s) + \gamma * |d|  \label{eq:loss_opl_gamma}
\end{align}
where $\gamma$ is the hyper-parameter controlling the weight for the two different constraints.

Since OPL acts only on intermediate features, we apply cross entropy loss over the outputs of the final classifier $\mathcal{H}_{\theta}$. The overall loss used  is a weighted combination of CE and OPL. We note that our proposed loss can also be used together with other common image classification losses, such as Guided Cross Entropy, Label Smoothing or even task specific loss functions in different computer vision tasks. The overall loss $\mathcal{L}$ can be defined as:
\begin{equation}
\mathcal{L} = \mathcal{L}_{\texttt{CE}} + \lambda \cdot \mathcal{L}_{\texttt{OPL}} \label{eq:loss_overall}
\end{equation}
where $\lambda$ is a hyper-parameter controlling the OPL weight.

\begin{algorithm}[tb]
   \caption{Pytorch style pseudocode for OPL}
   \label{alg:opl}
   
    \definecolor{codeblue}{rgb}{0.25,0.5,0.5}
    \lstset{
      basicstyle=\fontsize{7.2pt}{7.2pt}\ttfamily\bfseries,
      commentstyle=\fontsize{7.2pt}{7.2pt}\color{codeblue},
      keywordstyle=\fontsize{7.2pt}{7.2pt},
    }
    
\begin{lstlisting}[language=python] 
def forward(features, labels):
    """ 
    features:   features shaped (B, D) 
    labels:     targets shaped (B, 1)
    """
    features = F.normalize(features, p=2, dim=1)
    
    # masks for same and diff class features
    mask = torch.eq(labels, labels.t())
    eye = torch.eye(mask.shape[0])
    mask_pos = mask.masked_fill(eye, 0)
    mask_neg = 1 - mask
    
    # s & d calculation
    dot_prod = torch.matmul(features, features.t())
    pos_total = (mask_pos * dot_prod).sum()
    neg_total = torch.abs(mask_neg * dot_prod).sum()
    pos_mean =  pos_total / (mask_pos.sum() + 1e-6)
    neg_mean = neg_total / (mask_neg.sum() + 1e-6)
    
    # total loss
    loss = (1.0 - pos_mean) + neg_mean 
    
    return loss
\end{lstlisting}
\end{algorithm}

\begin{figure*}
    \centering
    \begin{subfigure}[b]{0.33\linewidth}        
        \centering
        \includegraphics[width=\linewidth]{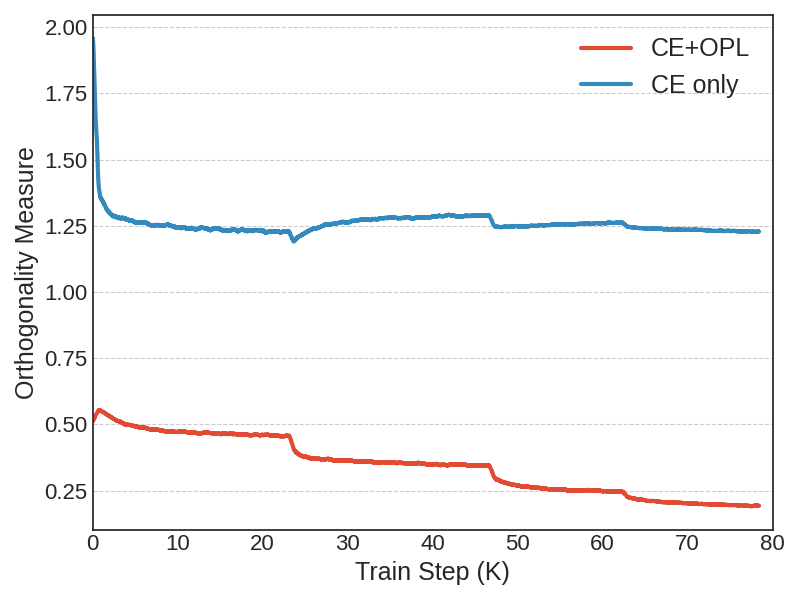}
        \caption{Feature orthogonality ($\downarrow$)}
    \end{subfigure}
    \begin{subfigure}[b]{0.33\linewidth}        
        \centering
        \includegraphics[width=\linewidth]{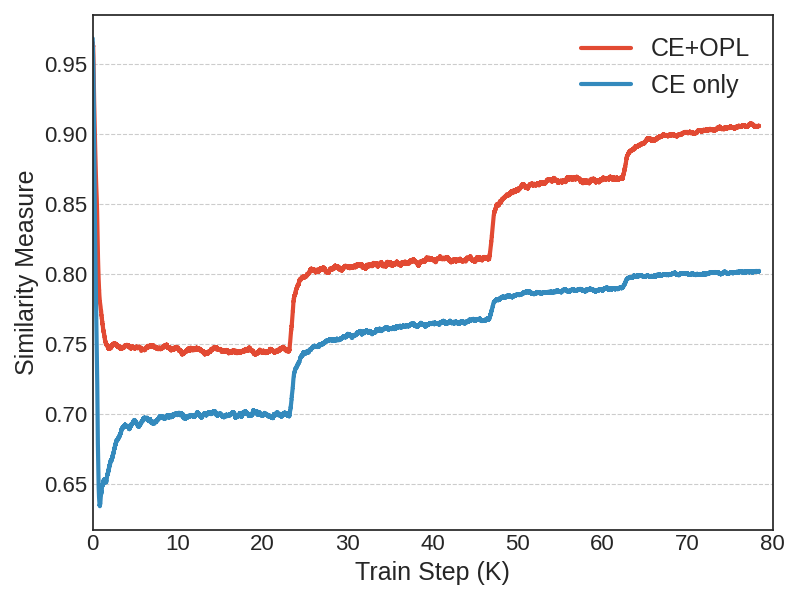}
        \caption{Similarity of same class features ($\uparrow$)}
    \end{subfigure}
    \begin{subfigure}[b]{0.33\linewidth}        
        \centering
        \includegraphics[width=\linewidth]{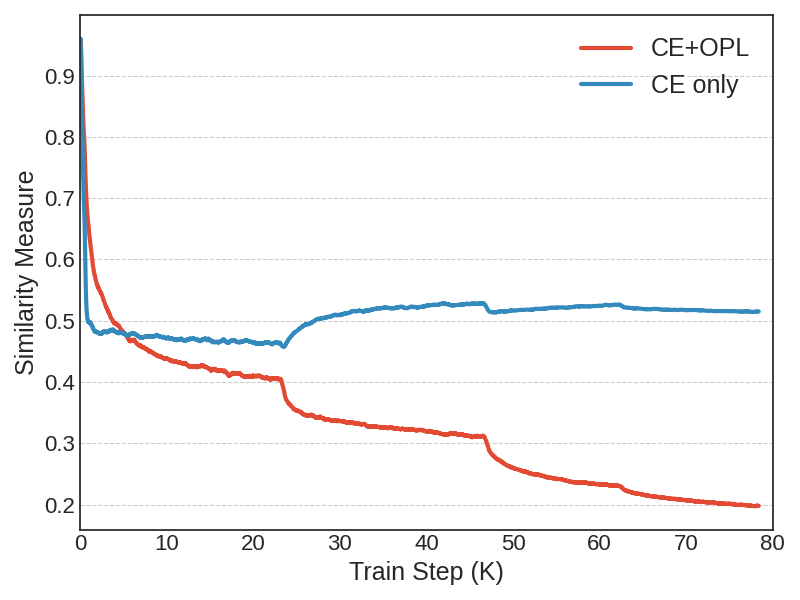}
        \caption{Similarity of different class features ($\downarrow$)}
    \end{subfigure} \vspace{-1.5em}
    	\caption{\textbf{Feature Analysis:} We compare feature orthogonality as measured by OPL and feature similarity as measured by cosine similarity and plot their convergence during training. Feature similarity is initially high because all features are random immediately after initialization. OPL simultaneously enforces higher inter-class similarity and intra-class dissimilarity in comparison with the CE baseline.}
	\vspace{-1.2em}
	\label{fig:ablation_train_plots}
\end{figure*}

\begin{figure}
    \centering
    \begin{subfigure}[b]{0.48\linewidth}        
        \centering
        \includegraphics[width=\linewidth]{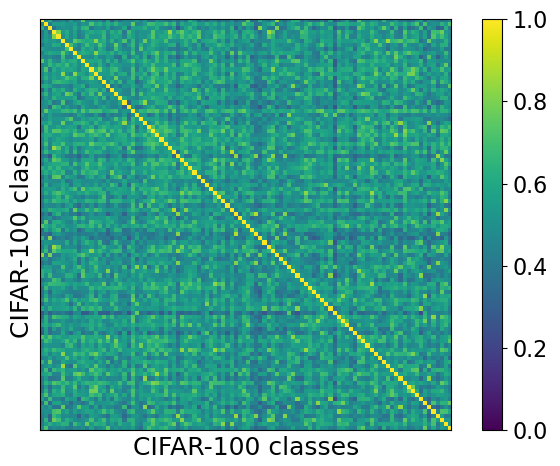}
        \caption{`CE-only' trained features}
        \label{fig:fig:ablation_block_plots_A}
    \end{subfigure}
    \begin{subfigure}[b]{0.48\linewidth}        
        \centering
        \includegraphics[width=\linewidth]{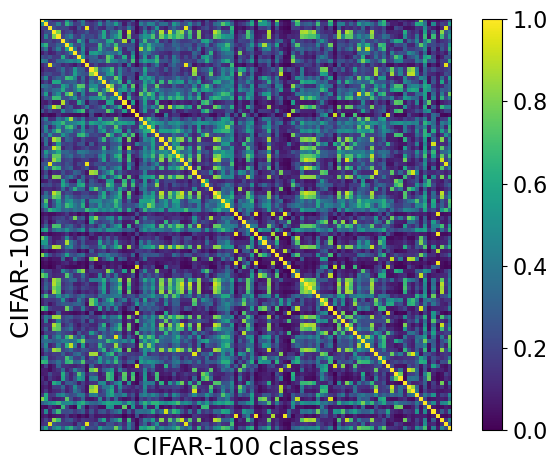}
        \caption{`CE+OPL' trained features}
        \label{fig:ablation_block_plots_B}
    \end{subfigure} \vspace{-0.5em}
    \caption{\textbf{Orthogonality Visualization:} We present matrices illustrating the orthogonality of average per class features computed over the CIFAR-100 test set. See \ref{sec: block_matrix_analysis} for more analysis.}
	\label{fig:ablation_block_plots} \vspace{-1.5em}
\end{figure}

\subsection{Interpretation and Analysis}
\textbf{Overall Objective:} Consider $\mathbf{F}_c$ is a set of mini-batch samples comprising of normalized features from the same class $c$ in a given dataset $\mathcal{D}$. The overall  OPL constraints can be viewed as a minimization of the following objective to update the network $\mathcal{H}_{\phi}$ over the random variables $\mathbf{F}_c$: 
\begin{align}
     \min_{\phi} \sum_{i=1}^{c} \sum_{j=1}^{c} \left| \mathop{\mathbb{E}}_{\scriptscriptstyle \substack{ (\mathbf{F}_i, \mathbf{F}_j) \sim \mathcal{D}}} \left[ \mathbf{F}_i \mathbf{F}_j^T \right] - \llbracket i=j \rrbracket \right|,
    \label{eq:opl_em}
\end{align}
where $| \cdot |$ is the absolute value operator and $\llbracket \cdot \rrbracket$ is the Iverson bracket operator. We refer to the term defined in Eq. \ref{eq:opl_em}  as the expected inter-class orthogonality. The behaviour of OPL in terms of minimizing this expectation is visualized in Fig. \ref{fig:ablation_block_plots} where average per-class feature vectors over the CIFAR-100 dataset are calculated for ResNet-56 models trained under `CE-only' and `CE+OPL' settings. Clear improvements over the CE baseline in terms of minimizing the expected inter-class orthogonality can be observed. Moreover, the stochastic mini-batch based application of OPL prevents naively pushing all non-diagonal values to zero as observed. This translates to allowing necessary inter-class relationships not encoded in one-hot labels of a dataset to be captured within the learned features.

\textbf{Decomposing OPL: }
Taking a step further, we decompose OPL into its sub-components, $s$ and $d$, as defined in Eq.~\ref{eq:loss_opl}. While $s$ computes the pair-wise cosine similarity between all same-class features within the mini-batch, $d$ calculates the similarity between different class features. These measures can directly be adopted to quantify the inter-class separation and intra-class compactness in any given features space. Moreover, the unweighted OPL formulation in Eq.~\ref{eq:loss_opl} can be considered a measure of the overall feature orthogonality 
within any given embedding space. It will be interesting to compare the contribution of OPL towards inter-class separation and intra-class clustering of a feature space in contrast to the generic CE based training scenario. We present this comparison by training ResNet-56 on CIFAR-100 dataset in Fig.~\ref{fig:ablation_train_plots}. This separation of features achieved through OPL translates to performance improvements not only within the standard classification setting, but also in tasks requiring transferable or generalizable features. Goldblum \etal \cite{unravelling_meta} explore the significance of inter-class separation and intra-class clustering for better performance when transferring a feature embedding to a few-shot learning task. Similar notions regarding discriminative features are explored in \cite{dg_class_margins} for domain generalization. We explore the effects of OPL in few-shot learning settings, and visualize the novel class embeddings learned with OPL in \ref{sec: vis_class_embeddings} using LDA 
\cite{lda_vis} to preserve the inter-class to intra-class variance ratio as suggested in \cite{unravelling_meta}.

\textbf{Why orthogonality constraints?:}
One may wonder what benefit orthogonality in the feature space can provide in comparison to simply maximizing margin between classes. Our reasoning is twofold: reducing sensitivity to batch composition and avoiding negative correlation constraints. Within the random mini-batch based training setting, the orthogonality objective provides a definitive geometric structure irrespective of the batch composition, while the optimal max-margin separation is dependent on the batch composition. Furthermore, in the common case where the output space feature dimension $d > c$ ($c$ number of classes), maximizing angular margin between normalized features on a unit hyper-sphere will lead to \emph{negative} correlation among the class prototypes (considering a maximal and equi-angular separation). We argue that this is an undesired constraint since the categorical classification task itself assumes non-existence of ordinal relationships between classes (\eg. use of orthogonal one-hot encoded labels). Moreover, extending the constraints to additionally cause negative correlation between classes unnecessarily focuses on already well-separated classes during training whereas our constraint which tends to ensure independence provides a more balanced objective to disentangle the class-specific characteristics of even more fine-grained classes.

\section{Experiments}
We extensively evaluate our proposed loss function on multiple tasks including \emph{image classification} (Tables \ref{table:comparison_cifar100} \& \ref{table:baseline_imagenet}), \emph{robustness against label noise} (Table \ref{table:noisy_cifar100}), \emph{robustness against adversarial attacks} (Table \ref{table:adversarial_robutness}) and \emph{generalization to domain shifts} (Table \ref{table:comparison_rsc}). We further observe the enhanced \emph{transferability} of orthogonal features \eg in the case of few shot learning (Tables \ref{table:comparison_rfs} \& \ref{table:comparison_metadataset}). Our approach shows consistent improvements and highlights the advantages of orthogonal features on this diverse set of tasks and datasets with various deep network backbones. Additionally, we demonstrate the plug-and-play nature of OPL by showing benefits of its use over CE, Truncated Loss (for noisy labels) \cite{truncated_loss}, RSC \cite{RSC_paper} and various adversarial learning baselines.

\subsection{Image Classification}
We evaluate the effectiveness of orthogonal features in the penultimate layer on image classification using our proposed training objective (Eq.~\ref{eq:loss_overall}). 
Competitive results are achieved showing consistent improvements  (Tables~\ref{table:comparison_cifar100} \&~\ref{table:baseline_imagenet}) on two datasets: CIFAR-100 \cite{krizhevsky2009learning}, and ImageNet \cite{krizhevsky2012imagenet}. 

\noindent\textbf{CIFAR-100} consists of 60,000 natural images spread across 100 classes, with 600 images per class. We apply OPL over a cross-entropy baseline for supervised classification on CIFAR-100 (following the experiment setup in \cite{negative_cam}), and compare our results against other loss functions which impose margin constraints \cite{sphere_face,large_margin_softmax,cos_face,arc_Face}, introduce regularization \cite{negative_cam,lezama2018ole,focal_loss,CB_focal_loss}, or promote clustering \cite{centre_loss,rbf_loss} to enhance separation among classes in Table \ref{table:comparison_cifar100}.  Despite its simplicity, our method performs well against state-of-the-art loss functions. Note that HNC \cite{negative_cam} is dependant on class activation maps, RBF \cite{rbf_loss} and LGM \cite{LGM_loss} involve learnable parameters, and CB Focal Loss \cite{CB_focal_loss} specifically solves class-imbalance. In contrast, OPL has a simple formulation easily integratable to any network architecture, involves no learnable parameters, and targets general classification. Additionally, we note how OPL has higher performance gains with respect to top-1 accuracy (in comparison to top-5 accuracy) which is the more challenging metric. We attribute this to the fact that increased separation through OPL mostly helps in classifying difficult samples. Further, we note top-5 is not a preferred measure for CIFAR-100 since most classes are different in nature as opposed to e.g., ImageNet with several closely related classes (where our gain is much pronounced for top-5 accuracy, as discussed next). 
\begin{table}[!htp]
\begin{center}
\setlength{\tabcolsep}{3pt}
\scalebox{0.88}{
\begin{tabular}{l|c|c|c|c}
\toprule[0.4mm]

\rowcolor{mygray} 
\cellcolor{mygray} &
  \multicolumn{2}{c|}{\cellcolor{mygray}Resnet-56} &
  \multicolumn{2}{c}{\cellcolor{mygray}ResNet-110} \\ \cline{2-5} 
\rowcolor{mygray} 
\multirow{-2}{*}{\cellcolor{mygray}Loss} &
  Top-1 &
  Top-5 &
  Top-1 &
  Top-5 \\ \midrule

Center Loss \venue{(ECCV'16)} \cite{centre_loss}      &72.72\%&93.06\%&74.27\%&93.20\%\\ 
Focal Loss \venue{(ICCV'17)} \cite{focal_loss}        &73.09\%&93.07\%&74.34\%&93.34\%\\ 
A-Softmax  \venue{(CVPR'17)} \cite{sphere_face}   &72.20\%&91.28\%&72.72\%&90.41\%\\ 
LMC Loss \venue{(CVPR'17)} \cite{cos_face}            &71.52\%&91.64\%&73.15\%&91.88\%\\ 
OLE Loss \venue{(CVPR'18)} \cite{lezama2018ole}       &71.95\%&92.52\%&72.70\%&92.63\%\\ 
LGM Loss \venue{(CVPR'18)} \cite{LGM_loss}            &73.08\%&93.10\%&74.34\%&93.06\%\\ 
Anchor Loss \venue{(ICCV'19)} \cite{anchor_loss}      &   -   &   -   &74.38\%&92.45\%\\ 
AAM Loss \venue{(CVPR'19)} \cite{arc_Face}            &71.41\%&91.66\%&73.72\%&91.86\%\\ 
CB Focal Loss \venue{(CVPR'19)} \cite{CB_focal_loss}  &73.09\%&93.07\%&74.34\%&93.34\%\\ 
HNC \venue{(ECCV'20)} \cite{negative_cam}             &73.47\%&\textbf{93.29\%}&74.76\%&\textbf{93.65\%}\\ 
RBF \venue{(ECCV'20)} \cite{rbf_loss}                 &73.36\%&92.94\%&   -   &   -   \\ \midrule
CE (Baseline)                                         &72.40\%&92.68\%&73.79\%&93.11\%\\ 
CE+OPL (Ours)                                         &\textbf{73.52\%}&93.07\%&\textbf{74.85\%}&93.32\%\\ \bottomrule[0.4mm]
\end{tabular}
}
\end{center}
\caption{\textbf{CIFAR-100}: These results indicate that a simple combination of cross-entropy along with our proposed orthogonal constraint gives improvements over the baseline loss function.}
\label{table:comparison_cifar100}
\end{table}

\begin{figure}[t]
    \centering
    \includegraphics[width=0.24\linewidth]{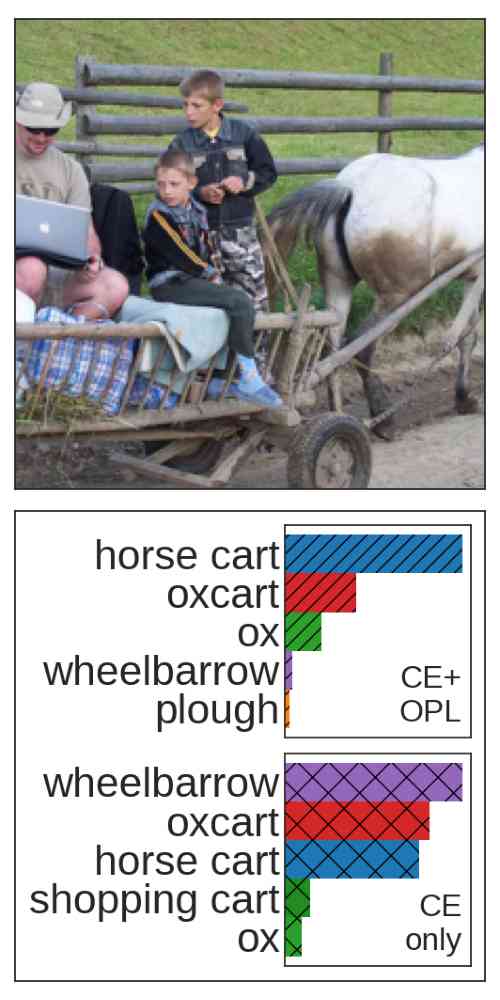}
    \includegraphics[width=0.24\linewidth]{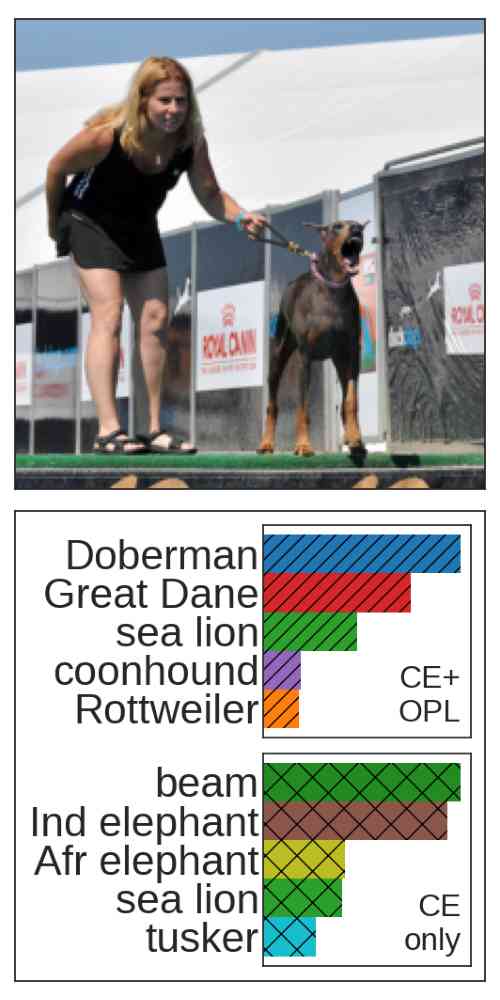}
    \includegraphics[width=0.24\linewidth]{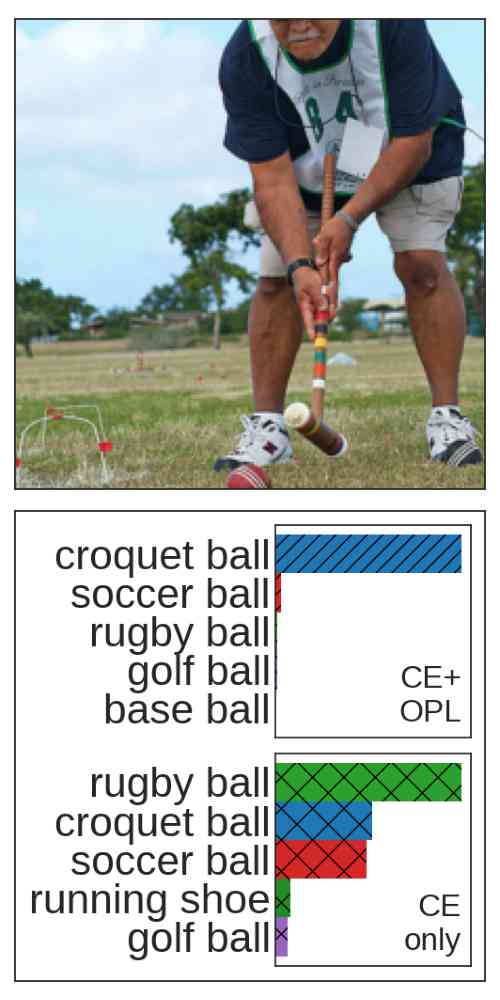}
    \includegraphics[width=0.24\linewidth]{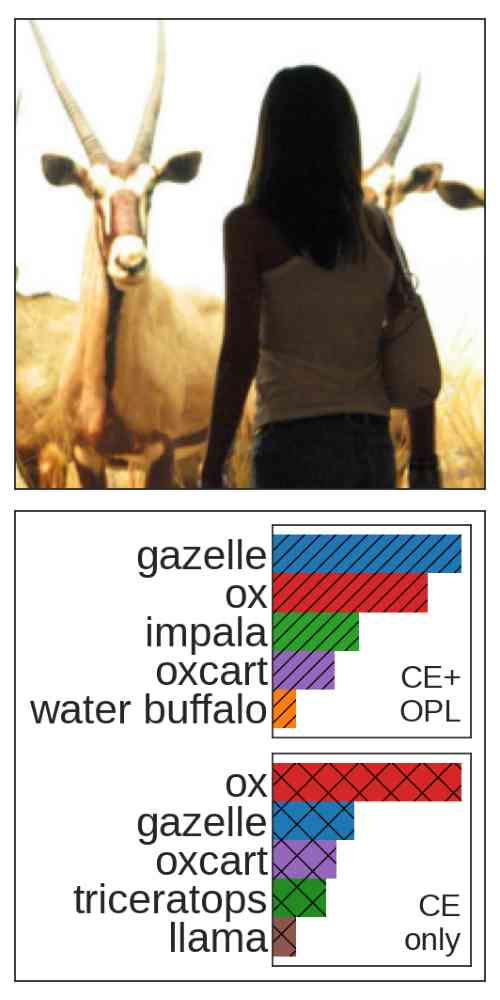}
    \caption{\textbf{Qualitative Results:} We present the top-5 predictions for OPL and CE in images where training with OPL has fixed the incorrect prediction by `CE only' model. See \ref{sec: imagenet_examples}.} \vspace{-1em}
    \label{fig:images_imnet}
\end{figure}

\noindent\textbf{ImageNet} is a standard large-scale dataset used in visual recognition tasks, containing roughly 1.2 million training images and 50,000 validation images. We experiment with OPL by integrating it to common backbone architectures used in image classification tasks: ResNet18 and ResNet50. We train \footnote{https://github.com/pytorch/examples/tree/master/imagenet} the models for 90 epochs using SGD with momentum (initial learning rate 0.1 decayed by 10 every 30 epochs). Results for these experiments are presented in Table \ref{table:baseline_imagenet} and Fig.~\ref{fig:images_imnet}. We note that simply enforcing our orthogonality constraints increases the top-1 (\%) accuracy of ResNet50 from 76.15\% to 76.98\% without any additional bells and whistles. Moreover, given the large number of fine-grained classes among the 1000 categories of ImageNet (\eg, multiple dog species) which can be viewed as difficult cases, the better discriminative features learned by OPL obtains notable improvements in top-5 accuracy as well.  
\begin{table}[!t]
	\small
	\begin{center}
\begin{tabular}{l|c|c|c|c}
\toprule[0.4mm]

\rowcolor{mygray} 
\cellcolor{mygray} & \multicolumn{2}{c|}{\cellcolor{mygray}ResNet-18}                 & \multicolumn{2}{c}{\cellcolor{mygray}ResNet-50}                 \\ \cline{2-5} 
\rowcolor{mygray} 
\multirow{-2}{*}{\cellcolor{mygray}Method} & top-1   & top-5   & top-1   & top-5   \\ \midrule

CE (Baseline)                     & 69.91\%           & 89.08\%           & 76.15\%           & 92.87\%           \\
CE + OPL (ours)         & \textbf{70.27\%}  & \textbf{89.60\%}  & \textbf{76.98\%}  & \textbf{93.30\%}  \\\bottomrule[0.4mm]
\end{tabular}
	\end{center}
	\caption{\textbf{Results on ImageNet:} OPL gives an improvement over a cross-entropy (CE) baseline for common backbone architectures.}
	\label{table:baseline_imagenet}
\end{table}

\subsection{Robustness against Label Noise}
Given the rich representation capacity of deep neural networks, especially considering how most can even fit random labels or noise perfectly \cite{random_labels_fit}, errors in the sample labels pose a significant challenge for training. In most practical applications, label noise is almost impossible to avoid, in particular, when it comes to large-scale datasets requiring millions of human annotations. Multiple works \cite{Ghosh2017RobustLF, truncated_loss} explore modifications to common objective functions aimed at building robustness to label noise. Despite the explicit inter-class separation constraints on the feature space enforced by OPL, we argue that the random mini-batch based optimization exploited by OPL negates the effects of noisy labels. This hypothesis is supported by our experiments presented in Table \ref{table:noisy_cifar100} which show additional robustness of OPL against label noise. We simply integrate OPL over the approach followed in \cite{truncated_loss}, without any task-specific modifications. 
\begin{table}
	\small
	\begin{center}
    \begin{tabular}{l|l|c|c}
    \toprule[0.4mm]
\rowcolor{mygray}    Dataset  & Method                        & Uniform         & Class Dependent \\ \midrule
    CIFAR10  & TL \venue{(NeurIPS'18)} \cite{truncated_loss}       & 87.62\%         & 82.28\%  \\
             & TL\cite{truncated_loss} + OPL &\textbf{88.45\%} &\textbf{87.02\%} \\ \midrule
    CIFAR100 & TL \venue{(NeurIPS'18)} \cite{truncated_loss}       & 62.64\%         & 47.66\%  \\
             & TL\cite{truncated_loss} + OPL &\textbf{65.62\%} &\textbf{53.94\%} \\ \bottomrule[0.4mm]
    \end{tabular}
	\end{center}
	\caption{\textbf{Results on CIFAR-100 for Noisy Labels:} We explore the effect of noisy labels when training with OPL for image classification tasks. We use the method in \cite{truncated_loss} as a baseline comparison with 0.4 noise level and ResNet18 backbone.} \vspace{-1.5em}
	\label{table:noisy_cifar100}
\end{table}

\subsection{Robustness against Adversarial Attacks}
Adversarial attacks modify a given benign sample by adding adversarial noise such that the deep neural network is deceived \cite{szegedy2013intriguing}. Adversarial examples are out-of-distribution samples and remain a challenging problem to solve. Adversarial training \cite{madry2017towards} emerges as an effective defense where adversarial examples are generated and added into the training set. We enforced orthogonality on such adversarial examples in the feature space while optimizing the model weights and show our benefit on different adversarial training mechanisms \cite{madry2017towards, hendrycks2019using, Wang2020Improving}. Important to note that all the considered adversarial training schemes \cite{madry2017towards, hendrycks2019using, Wang2020Improving} are different in nature \eg  training in Madry \etal \cite{madry2017towards} is based on cross-entropy only, Hendrycks \etal \cite{hendrycks2019using} propose to exploit pre-training, while Wang \etal \cite{Wang2020Improving} introduce a surrogate loss along with cross-entropy. Our orthogonality constraint help maximizing adversarial robustness in all cases showing the generic and plug-and-play nature of our proposed loss. In order to have reliable evaluation,  we report robustness gains against Auto-Attack (AA) \cite{croce2020reliable} in Table \ref{table:adversarial_robutness}. On CIFAR10, our method increased robustness of \cite{madry2017towards} by $5.11\%$, \cite{hendrycks2019using} by $0.81\%$ and \cite{Wang2020Improving} by $2.2\%$.

\begin{table}[!t]
\begin{center}
	\small
	\scalebox{0.95}[0.95]{
	\begin{tabular}{l|l|c|c}
	\toprule[0.4mm]
\rowcolor{mygray} 	Dataset       & Method & Clean  & Advers.  \\ \midrule
	\multirow{6}{*}
	{CIFAR10}& Madry \etal \venue{(ICLR'18)} \cite{madry2017towards}    &  87.14 &  44.04    \\ 
	&   Madry \etal \cite{madry2017towards} + OPL & \textbf{87.76}& \textbf{49.15}\\
	\cline{2-4}
	& Hendrycks \etal \venue{(PMLR'19)} \cite{hendrycks2019using} & 87.11 & 54.92\\
	&  Hendrycks \etal \cite{hendrycks2019using} + OPL & \textbf{87.51} & \textbf{55.73}\\
	\cline{2-4}
	& MART \cite{Wang2020Improving} \venue{(ICLR'20)} & \textbf{84.49} & 54.10\\
	&  MART\cite{Wang2020Improving} + OPL&84.41 & \textbf{56.23}\\
	\midrule
	\multirow{6}{*}{CIFAR100}& Madry \etal \venue{(ICLR'18)} \cite{madry2017towards}    & 60.20 &  20.60    \\ 
	&   Madry \etal \cite{madry2017towards} + OPL &\textbf{61.13} &  \textbf{23.01} \\
	\cline{2-4}
	& Hendrycks \etal \venue{(PMLR'19)} \cite{hendrycks2019using} & 59.23	&	28.42\\
	&  Hendrycks \etal \cite{hendrycks2019using} + OPL &\textbf{61.00}&	\textbf{30.05} \\
	\cline{2-4}
	& MART \venue{(ICLR'20)} \cite{Wang2020Improving} & \textbf{58.90}& 23.40\\
	&  MART\cite{Wang2020Improving} + OPL&58.01 & \textbf{25.74}\\
	\bottomrule[0.4mm]
\end{tabular}}
\end{center}
\caption{\textbf{OPL performance on Adversarial Robustness}: We show the impact of enforcing orthogonality on the robust features. We adversarially train baseline methods \cite{madry2017towards, hendrycks2019using, Wang2020Improving} by adding OPL constraint during training.  Robust features obtained with OPL leads to better accuracy and show clear improvements over the baseline. Top-1 accuracy is reported against Auto-Attack \cite{croce2020reliable} in whitebox setting (attacker has full knowledge of the model architecture and pretrained weights).}
\label{table:adversarial_robutness}
\end{table}

\subsection{Domain Generalization (DG)}
The DG problem aims to train a model using multi-domain source data such that it can directly generalize to new domains without the need of retraining. We argue that the feature space constraints of OPL tend to capture more general semantic features in images which generalize better across domains. This is verified by the performance improvements for DG that we obtain by integrating OPL with the state-of-the-art approach in \cite{RSC_paper} and evaluating on the popular PACS dataset \cite{pacs_dataset}. The results presented in Table \ref{table:comparison_rsc} indicate that integrating OPL with \cite{RSC_paper} sets new state-of-the-art across all four domains as compared to \cite{RSC_paper}.

\begin{table}[]
	\small
\begin{center}\setlength{\tabcolsep}{3pt}
\begin{tabular}{l|c|c|c|c|c}
\toprule[0.4mm]
\rowcolor{mygray} Method                      & Art   & Cartoon & Sketch & Photo & Avg \\ \midrule
JiGen\venue{(CVPR'19)} \cite{jigen_paper}    & 86.20 & 78.70   & 70.63  & 97.66 & 83.29   \\ 
MASF\venue{(NeurIPS'19)} \cite{mafs_paper}      & 82.89 & 80.49   & 72.29  & 95.01 & 82.67   \\ 
MetaReg\venue{(NeurIPS'18)} \cite{metaReg_paper}& 87.20 & 79.20   & 70.30  & 97.60 & 83.60   \\ 
RSC\venue{(ECCV'20)} \cite{RSC_paper}        & 87.89 & 82.16   & 83.35  & 96.47*& 87.47   \\ \hline
RSC + OPL                        & \textbf{88.28} & \textbf{84.64}   & \textbf{84.17}  & \textbf{96.83} & \textbf{88.48}   \\ \bottomrule[0.4mm]
\end{tabular}
\end{center}
\caption{\textbf{Results on PACS dataset:} We integrate OPL with \cite{RSC_paper}, gaining improvements for domain generalization tasks (*best replicated value).}\vspace{-2em}
\label{table:comparison_rsc}
\end{table}

\begin{table*}[]
\begin{center} 
\resizebox{\textwidth}{!}{
	\begin{tabular}{l|c|c|c|c|c|c|c} 
	\toprule[0.4mm]
\rowcolor{mygray}	Method    & New Loss             & Cifar:1shot             & Cifar:5shot              & Mini:1shot              & Mini:5shot                & Tier:1shot                & Tier:5shot              \\ \midrule
	MAML\venue{(PMLR'17)} \cite{maml}    &  -  & 58.90$\pm$1.9           & 71.50$\pm$1.0            & 48.70$\pm$1.84          & 63.11$\pm$0.92            & 51.67$\pm$1.81            & 70.30$\pm$1.75          \\ 
	PN \venue{(NIPS'17)} \cite{prototypical_network} & -  & 55.50$\pm$0.7           & 72.00$\pm$0.6            & 49.42$\pm$0.78          & 68.20$\pm$0.66            & 53.31$\pm$0.89            & 72.69$\pm$0.74          \\ 
	RN\venue{(CVPR'18)} \cite{relation_network}   &  -  & 55.00$\pm$1.0           & 69.30$\pm$0.8            & 50.44$\pm$0.82          & 65.32$\pm$0.70            & 54.48$\pm$0.93            & 71.32$\pm$0.78          \\ 
	Shot-Free\venue{(ICCV'19)} \cite{shot_free}    &     -      & 69.20$\pm$N/A           & 84.70$\pm$N/A            & 59.04$\pm$N/A           & 77.64$\pm$N/A             & 63.52$\pm$N/A             & 82.59$\pm$N/A           \\ 
	MetaOptNet\venue{(CVPR'19)} \cite{meta_opt_net}        &    -  & 72.60$\pm$0.7           & 84.30$\pm$0.5            & 62.64$\pm$0.61          & 78.63$\pm$0.46            & 65.99$\pm$0.72            & 81.56$\pm$0.53          \\ \hline
	RFS\venue{(ECCV'20)}\cite{rfs_paper}   & - & 71.45$\pm$0.8           & 85.95$\pm$0.5            & 62.02$\pm$0.60          & 79.64$\pm$0.44            & 69.74$\pm$0.72            & 84.41$\pm$0.55          \\ 
	\textbf{RFS + OPL (Ours)}   & \checkmark     & \textbf{73.02$\pm$0.4}  & \textbf{86.12$\pm$0.2}   & \textbf{63.10$\pm$0.36} & \textbf{79.87$\pm$0.26}   & \textbf{70.20$\pm$0.41}   & \textbf{85.01$\pm$0.27} \\ \hline \hline
	NAML\venue{(CVPR'20)} \cite{aml_paper}   & \checkmark  & -                       & -                        & 65.42$\pm$0.25          & 75.48$\pm$0.34            & -                         & -                       \\ 
	Neg-Cosine\venue{(ECCV'20)} \cite{negative_margin}& \checkmark  &  -               & -                        & 63.85$\pm$0.81          & 81.57$\pm$0.56            & -                         & -                       \\ \hline
	SKD\venue{(Arxiv'20)} \cite{skd_paper} & \checkmark     & 74.50$\pm$0.9           & 88.00$\pm$0.6            & 65.93$\pm$0.81          & 83.15$\pm$0.54            & 71.69$\pm$0.91            & 86.66$\pm$0.60          \\ 
	\textbf{SKD + OPL (Ours)} & \checkmark       & \textbf{74.94$\pm$0.4}  & \textbf{88.06$\pm$0.3}   & \textbf{66.90$\pm$0.37} & \textbf{83.23$\pm$0.25}   & \textbf{72.10$\pm$0.41}   & \textbf{86.70$\pm$0.27} \\ \bottomrule[0.4mm]
\end{tabular}}
\end{center}
\caption{\textbf{Few-Shot Learning Improvements:} We obtain performance improvements using OPL over the RFS \cite{rfs_paper} baseline and SKD baseline \cite{skd_paper} containing ResNet-12 backbones. Our loss is simply plugged in to their supervised feature learning phase. Results reported for our experiment are averaged over 3000 episodic runs. Note that \cite{skd_paper,aml_paper,negative_margin} are recent loss functions specific to FSL.}
\label{table:comparison_rfs}
\end{table*}

\begin{table}[]
\begin{center}
\setlength{\tabcolsep}{7pt}
\scalebox{0.95}{
	\begin{tabular}{l|c|c|c}
	\toprule[0.4mm]
\rowcolor{mygray}	Dataset                 & \begin{tabular}{@{}c@{}} CNAPs \cite{cnaps_paper} \\ \venue{(NeurIPS'19)} \end{tabular}  &  \begin{tabular}{@{}c@{}} SUR \cite{SUR_paper} \\ \venue{(ECCV'20)} \end{tabular}    & \begin{tabular}{@{}c@{}} SUR + OPL \\ (Ours) \end{tabular} \\ \midrule
	Imagenet                & 52.3$\pm$1.0              & 56.4$\pm$1.2              & \textbf{56.5$\pm$1.1}             \\ 
	Omniglot                & 88.4$\pm$0.7              & 88.5$\pm$0.8              & \textbf{89.8$\pm$0.7}             \\ 
	Aircraft                & \textbf{80.5$\pm$0.6}     & 79.5$\pm$0.8              & 79.6$\pm$0.7                      \\ 
	Birds                   & 72.2$\pm$0.9              & 76.4$\pm$0.9              & \textbf{76.9$\pm$0.7}             \\ 
	Textures                & 58.3$\pm$0.7              & \textbf{73.1$\pm$0.7}     & 72.7$\pm$0.7                      \\ 
	Quick Draw              & 72.5$\pm$0.8              & 75.7$\pm$0.7              & \textbf{75.7$\pm$0.7}             \\ 
	Fungi                   & 47.4$\pm$1.0              & 48.2$\pm$0.9              & \textbf{50.1$\pm$1.0}             \\ 
	VGG Flower              & 86.0$\pm$0.5              & 90.6$\pm$0.5              & \textbf{90.9$\pm$0.5}             \\ 
	MSCOCO                  & 42.6$\pm$1.1              & \textbf{52.1$\pm$1.0}     & 52.0$\pm$1.0                      \\ 
	MNIST                   & 92.7$\pm$0.4              & 93.2$\pm$0.4              & \textbf{94.3$\pm$0.4}             \\ 
	CIFAR10                 & 61.5$\pm$0.7              & 66.4$\pm$0.8              & \textbf{66.6$\pm$0.7}             \\ 
	CIFAR100                & 50.1$\pm$1.0              & 57.1$\pm$1.0              & \textbf{57.6$\pm$1.0}             \\ \hline
	Average                 & 67.0                      & 71.4                      & \textbf{71.9}                     \\ \bottomrule[0.4mm]
\end{tabular}}
\end{center}
\caption{\textbf{Results on Meta-Dataset:} OPL is integrated with the SUR-PNF method in \cite{SUR_paper} for the Meta-Dataset train on all setting. \textit{Traffic Signs} dataset has been omitted in comparisons due to an error in Meta-Dataset possibly affecting prior work.}
\label{table:comparison_metadataset}
\end{table}

\begin{table}[!t]
\begin{center}\setlength{\tabcolsep}{10pt}
\scalebox{0.9}{
\begin{tabular}{l|c|c|c}
\toprule[0.4mm]
\rowcolor{mygray} hyper-parameter      & $\gamma$=2             & $\gamma$=1       & $\gamma$=0.5   \\ \midrule
$\lambda$ = 0.05            & 70.48 & 70.66 & 72.02 \\ 
$\lambda$ = 0.1             & 70.12 & 70.94 & 71.30 \\ 
$\lambda$ = 0.5             & 70.26 & 71.18 & 70.66 \\ 
$\lambda$ = 1               & 69.78 & 70.48 & \textbf{72.20} \\ 
$\lambda$ = 2               & 67.64 & 69.58 & 70.52 \\ \bottomrule[0.4mm]
\end{tabular}}
\end{center}
\caption{\textbf{Hyper-parameter search:} We report the top-1 accuracy values on a held-out validation set on CIFAR-100 using ResNet-56 backbone after training using OPL with various pairs of $\lambda$ and $\gamma$ hyper-parameters.}\vspace{-1em}
\label{table:search_table}
\end{table}

\subsection{Few Shot Learning (FSL)}
In this section, we explore the transferability of features learned with our loss function in relation to FSL tasks. We evaluate OPL on three benchmark few-shot classification datasets: miniImageNet, tieredImageNet, and CIFAR-FS. We run additional experiments on Meta-Dataset \cite{meta_dataset} which is a large-scale benchmark for evaluating FSL methods in more diverse and challenging settings. Similar to \cite{cnaps_paper}, we expand Meta-Dataset by adding three additional datasets, MNIST, CIFAR10, and CIFAR100. 
In light of work that shows promise of  learning strong features for FSL
\cite{rfs_paper}, we experiment with OPL using it as an auxiliary loss on the feature space during the supervised training. Quantitative results highlighting the performance improvements are presented in Table \ref{table:comparison_rfs}. Our results on Meta-Dataset are obtained using the \textit{train on all} setting presented in \cite{meta_dataset}. We integrate OPL over the method presented in \cite{SUR_paper}, train on the first 8 datasets of Meta-Dataset, and evaluate on the rest (including the three additional datasets from \cite{cnaps_paper}). Results are presented in Table \ref{table:comparison_metadataset}.  See \ref{sec: robustness_to_noise_fsl} for robustness of OPL features against \emph{sample noise} in FSL tasks.

\begin{figure}[!t]
    \centering
    \begin{subfigure}[b]{0.48\linewidth}        
        \centering
        \includegraphics[width=\linewidth]{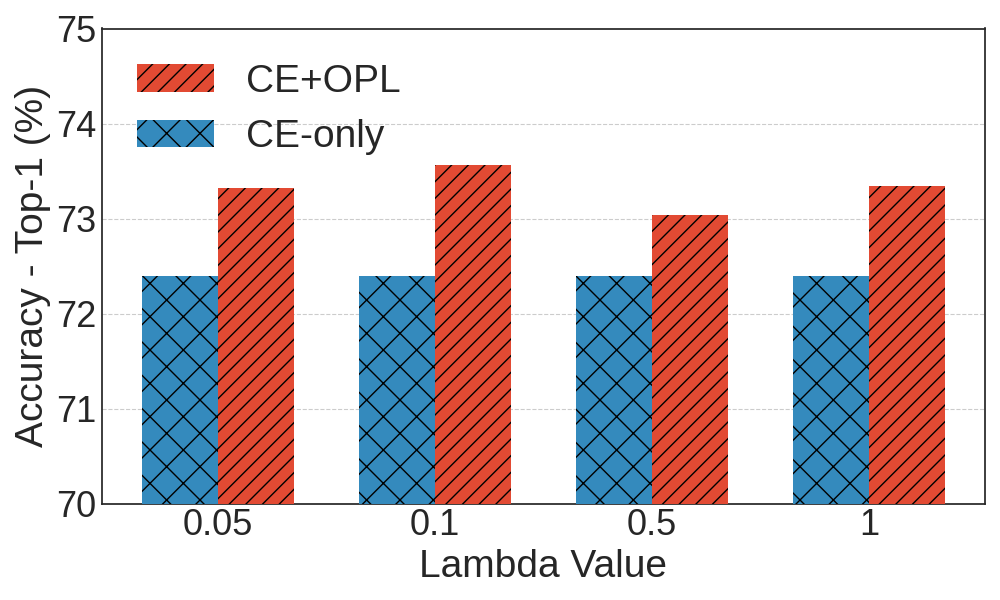}
    	\caption{Sensitivity to $\lambda$}
    	\label{fig:ablation_lambda}
    \end{subfigure}
    \begin{subfigure}[b]{0.48\linewidth}        
        \centering
        \includegraphics[width=\linewidth]{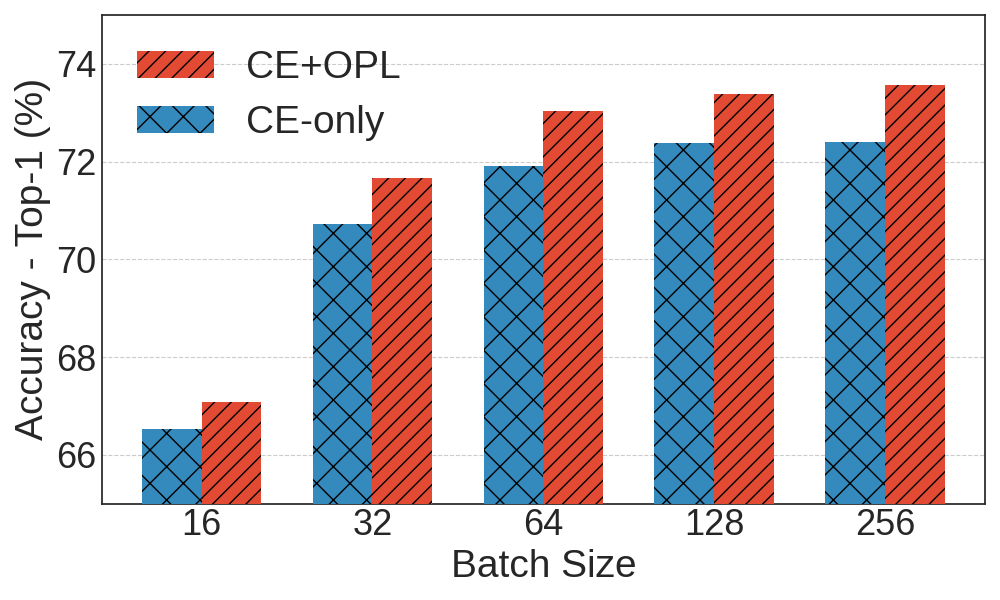}
    	\caption{Variation with Batch Size}
    	\label{fig:ablation_batch_size}
    \end{subfigure}
    \caption{\textbf{Ablative Study:} OPL achieves consistent performance improvements against a CE-only baseline when evaluated on CIFAR-100 dataset with a ResNet-56 backbone.}\vspace{-1em}
\end{figure}

\subsection{Ablative Study}
OPL in its full form (Eq.~\ref{eq:loss_opl_gamma}) contains two hyper-parameters, $\lambda$ and $\gamma$. We conduct a hyper-parameter search over a held-out validation set of CIFAR-100  (see Table~\ref{table:search_table}). The optimum values selected from these experiments are kept consistent across all other tasks and used when reporting test performance. Furthermore, we evaluate the performance of OPL on the test split of CIFAR-100 for varying $\lambda$ values keeping $\gamma$ fixed, to illustrate the minimal sensitivity of our method to different $\lambda$ values in Fig.~\ref{fig:ablation_lambda}. 

Next, we consider how OPL operates on random mini-batches, and evaluate its performance against varying batch sizes (CIFAR-100 dataset). These results presented in Fig. \ref{fig:ablation_batch_size} exhibit how OPL consistently 
\section{Conclusion}

We present a simple yet effective loss function to enforce orthogonality on the output feature space and establish its performance improvements for a wide range of classification tasks. Our loss function operates in conjunction with the softmax CE loss, and can be easily integrated with any DNN. We also explore a variety of characteristics of the features learned with OPL illustrating its benefit for few-shot learning, domain-generalization and robustness against adversarial attacks and label noise. In future, we hope to explore other variants of OPL including its adaptation to unsupervised representation learning.

{\small
\bibliographystyle{ieee_fullname}
\bibliography{egbib}
}

\newpage\leavevmode\thispagestyle{empty}\newpage \clearpage
\setcounter{section}{0}
\setcounter{table}{0}
\setcounter{figure}{0}
\def\thesection{Appendix \Alph{section}}
\twocolumn[
  \begin{@twocolumnfalse}
    \begin{center}
        \section*{\Large{Supplementary: Orthogonal Projection Loss}}
        \vspace{1.5em}
    \end{center}
  \end{@twocolumnfalse}
]

In this supplementary document we include: 
\begin{itemize}\setlength{\itemsep}{0em}
    \item[--] Additional comparisons with the baseline on the MNIST dataset (\ref{sec:mnist}). 
    \item[--] OPL performance with scalable neural architecture method \cite{Neural_Scale} (\ref{sec:scalable}).
    \item[--] Robustness of OPL against noise in the input images (\ref{sec: robustness_to_noise_fsl}).
    \item[--] Visualization of classification results (\ref{sec:visualization}).
\end{itemize}
\section{Experimentation}
Here, we present results of experiments conducted using OPL on a set of additional settings.

\subsection{Digit Classification (MNIST)}\label{sec:mnist}
We conduct experiments on the MNIST dataset integrating OPL over a CE baseline. We use a 4-layer convolutional neural network with 32-dimensional feature embedding (after a global average pool operation) following the experimental setup in \cite{rbf_loss}. Our results are reported in Table \ref{table:mnist_results}. Additionally, we conduct experiments appending a fully-connected layer to reduce the feature dimensionality to 2 for generating better visualizations on behaviour of OPL in feature-space (presented in Fig. \ref{fig:opl_explanation} in main article). 
\begin{table}[h]
\resizebox{\columnwidth}{!}{
\begin{tabular}{l|c|c|c|c}
\toprule[0.4mm]
\rowcolor{mygray} Method  & 1st     & 2nd     & 3rd     & Avg     \\ \midrule
CE (baseline)      & 99.28\% & 99.27\% & 99.25\% & 99.27\% \\ 
CE+OPL (ours)     & 99.58\% & 99.56\% & 99.61\% & 99.58\% \\ \bottomrule[0.4mm]
\end{tabular}}
\vspace{4mm}
\caption{\textbf{Results on MNIST:} OPL obtains improvements over the CE baseline on MNIST dataset. Each experiment is replicated thrice and the average across runs is also reported.}
\label{table:mnist_results}

\end{table}

\subsection{Scalable Architectures}\label{sec:scalable}
We consider the recent Neural Architecture Scaling approach  proposed in \cite{Neural_Scale} and plug-in our OPL on top of it to study our scalability.  Refer Table~\ref{table:comparison_cifar100_small} for the results. 
\begin{table}[H]
\small
\begin{center}
	\begin{tabular}{l|c|c|c}
	\toprule[0.4mm]
\rowcolor{mygray}	Method                         & Backbone   & Baseline\cite{Neural_Scale} & \cite{Neural_Scale} + OPL        \\ \midrule
	NeuralScale \cite{Neural_Scale}& ResNet18   & 77.59\%  & \textbf{77.81\%}  \\ \hline
	NeuralScale \cite{Neural_Scale}& VGG11     	& 67.42\%  & \textbf{67.69\%}  \\ \bottomrule[0.4mm]
\end{tabular}
\end{center}
\caption{\textbf{Additional results on CIFAR-100:} Performance improvements integrating OPL into small-scalable backbones for classification. Reported values are top-1 classification accuracies.}
\label{table:comparison_cifar100_small}
\end{table}

\subsection{Robustness to Noise: FSL}
\label{sec: robustness_to_noise_fsl}
We have already established through empirical evidence how OPL improves performance for few-shot learning tasks as well as robustness to adversarial examples present during evaluation. We now explore the more challenging task of exploring robustness to input sample noise in a FSL setting (similar to the one in \ref{sec: vis_class_embeddings}). The base training is conducted with no noise present in training data. During evaluation, the support and query set images are corrupted with random Gaussian noise of varying standard deviation (referred to as $\sigma$). This can be considered a domain shift on top of unseen novel classes during evaluation. The features learned with OPL during base training exhibit better robustness to such input corruptions in this FSL setting. We report these results in Table \ref{table:comparison_rfs_noise}. The experiments conducted followed the method in \cite{rfs_paper} integrated with OPL.

\begin{table*}[!ht]
\begin{center}
	\begin{tabular}{l|c|c|c|c|c|c|c}
	\toprule[0.4mm]
\rowcolor{mygray}	Method                  & Noise             & Cifar:1shot             & Cifar:5shot         & Mini:1shot          & Mini:5shot          & Tier:1shot          & Tier:5shot          \\ \midrule
	RFS \cite{rfs_paper}    &($\sigma=0.1$)     & 63.30$\pm$0.39          & 80.36$\pm$0.28          & 55.98$\pm$0.37          & 74.46$\pm$0.27          & 66.54$\pm$0.43          & 82.92$\pm$0.29          \\ \hline
	RFS\cite{rfs_paper} + OPL&($\sigma=0.1$)     & \textbf{65.42$\pm$0.40} & \textbf{81.41$\pm$0.30} & \textbf{56.21$\pm$0.36} & 73.20$\pm$0.29          & \textbf{66.60$\pm$0.41} & \textbf{83.21$\pm$0.29} \\ \hline \hline
	RFS \cite{rfs_paper}    &($\sigma=0.05$)    & 68.32$\pm$0.38          & 84.34$\pm$0.27          & 60.22$\pm$0.36          & 77.45$\pm$0.27          & 68.65$\pm$0.41          & 83.12$\pm$0.27          \\ \hline
	RFS\cite{rfs_paper} + OPL& ($\sigma=0.05$)   & \textbf{71.05$\pm$0.41} & \textbf{84.46$\pm$0.28} & \textbf{61.70$\pm$0.37} & \textbf{77.59$\pm$0.27} & \textbf{69.60$\pm$0.40} & \textbf{84.50$\pm$0.29} \\ \bottomrule[0.4mm]
	\end{tabular}
\end{center}
\caption{\textbf{Additional FSL Experiments:} We explore the robustness of models to noise (random Gaussian noise of varying standard deviation is added to input images) in FSL setting. Models trained with our proposed OPL loss are significantly more robust compared to the cross-entropy only baseline in \cite{rfs_paper}.}
\label{table:comparison_rfs_noise}

\end{table*}
\begin{figure}[!htb]
	\begin{center}
		\includegraphics[width=0.48\linewidth]{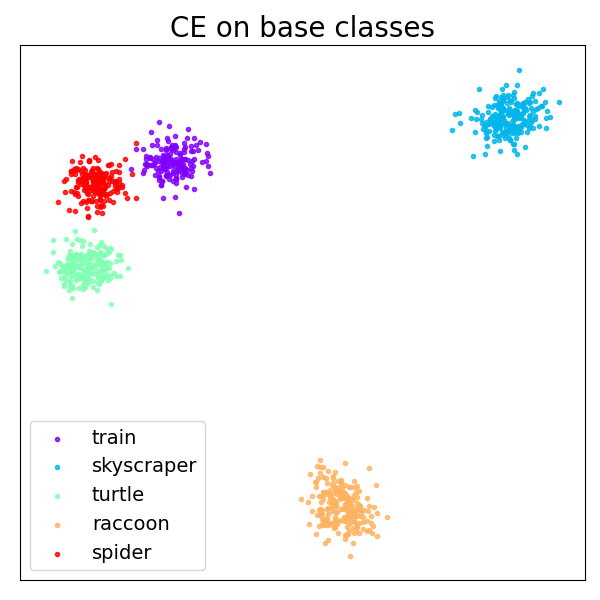}
		\includegraphics[width=0.48\linewidth]{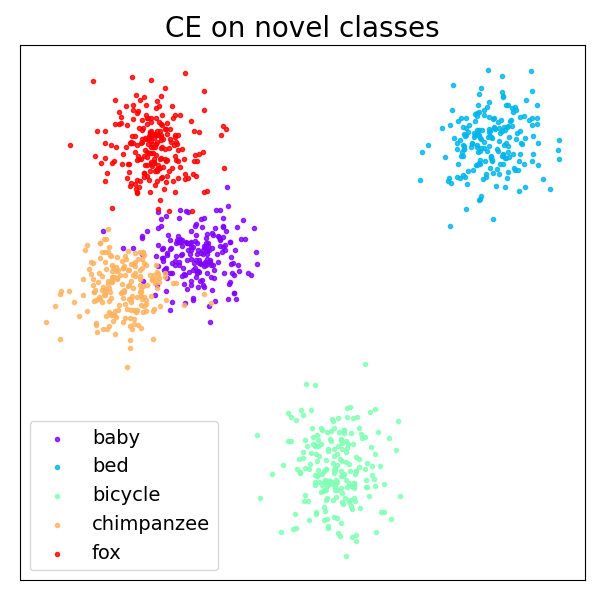}
		\includegraphics[width=0.48\linewidth]{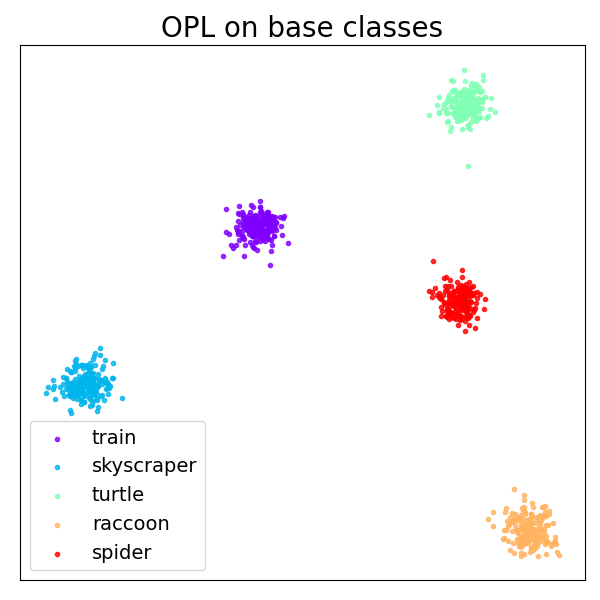}
		\includegraphics[width=0.48\linewidth]{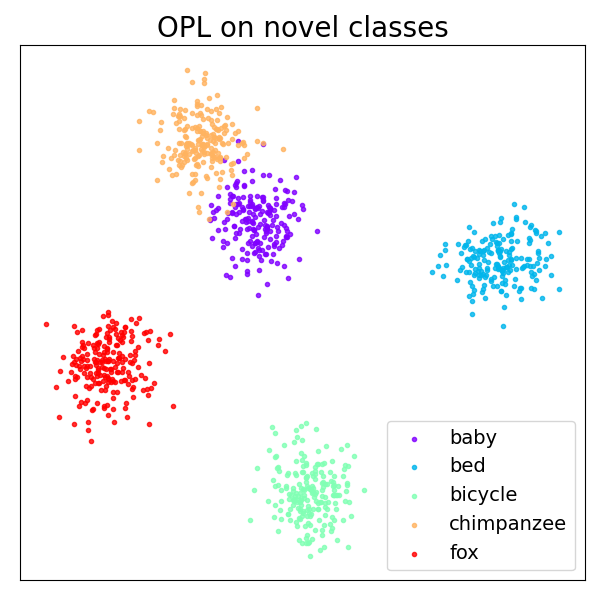}
	\end{center}
	\vspace{-4mm}
	\caption{\textbf{LDA visualization for CE vs OPL in FSL setting:} Training with OPL increases separation of features in both base and novel classes when applied in a few-shot learning setting. LDA has been used following the insights in \cite{unravelling_meta}. 
	}
	\label{fig:ce_vs_opl_fsl}
\end{figure}

\section{Visualization}\label{sec:visualization}
\begin{figure*}
    \centering
    \includegraphics[width=0.24\linewidth]{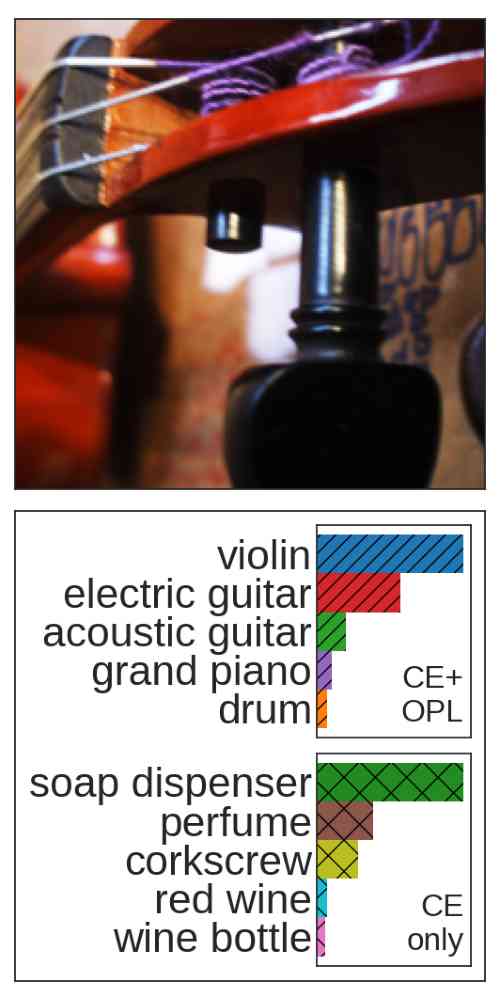}
    \includegraphics[width=0.24\linewidth]{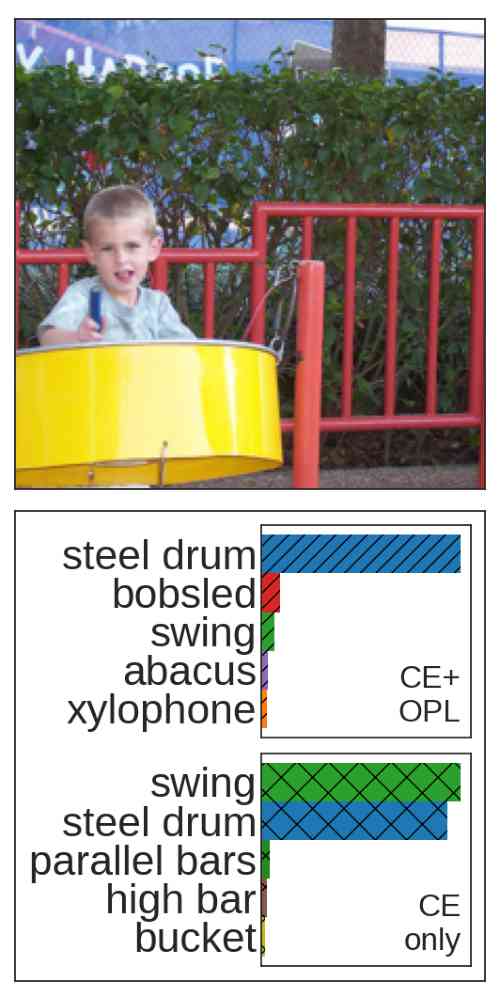}
    \includegraphics[width=0.24\linewidth]{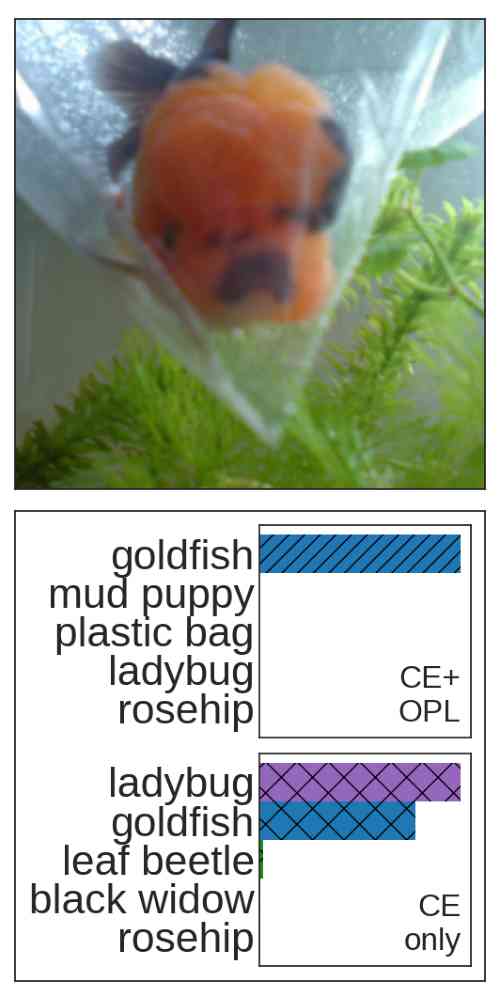}
    \includegraphics[width=0.24\linewidth]{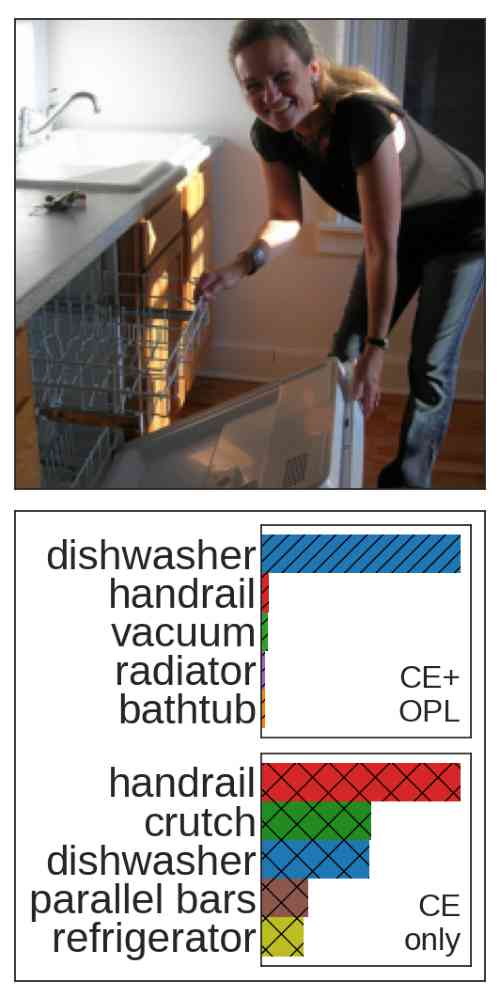}
    \caption{\textbf{Visualization of Images:} we show images where OPL predicts the correct but CE fails.}
    \label{fig:app_imnet_02}
\end{figure*}

In this section, we present additional visualizations exploring various aspects of OPL and its performance.

\subsection{Class Embeddings}\label{sec: vis_class_embeddings}
Consider a few-shot learning setting, where a model trained in a fully-supervised manner (referred to as base model / base training) on a set of selected classes which contains training labels (referred to as base classes) is later evaluated on a set of unseen classes (referred to as novel classes). The sets of base and novel classes are disjoint. The evaluation protocol would involve episodic iterations, where in each step a small set of labelled samples from the novel classes (referred to as support set) is available during inference, and an another set of those same novel classes (referred to as query set) is available for calculating the accuracy metrics. 

Given how our proposed loss is already able to explicitly enforce constraints on the feature space during base training, we want to examine if the additional discriminative nature endowed on the features by OPL is aware of higher level semantics. To evaluate this, we explore the more challenging task of inter-class separation and intra-class clustering of novel classes which are unseen during the base training. We train a model following the approach in \cite{rfs_paper} integrating OPL, and visualize the separation of different class features for both base and novel classes in Fig. \ref{fig:ce_vs_opl_fsl}. 

\subsection{ImageNet Examples}
\label{sec: imagenet_examples}
\begin{figure*}
    \centering
    \includegraphics[width=0.24\linewidth]{figures/plots/images/joint_008.jpg}
    \includegraphics[width=0.24\linewidth]{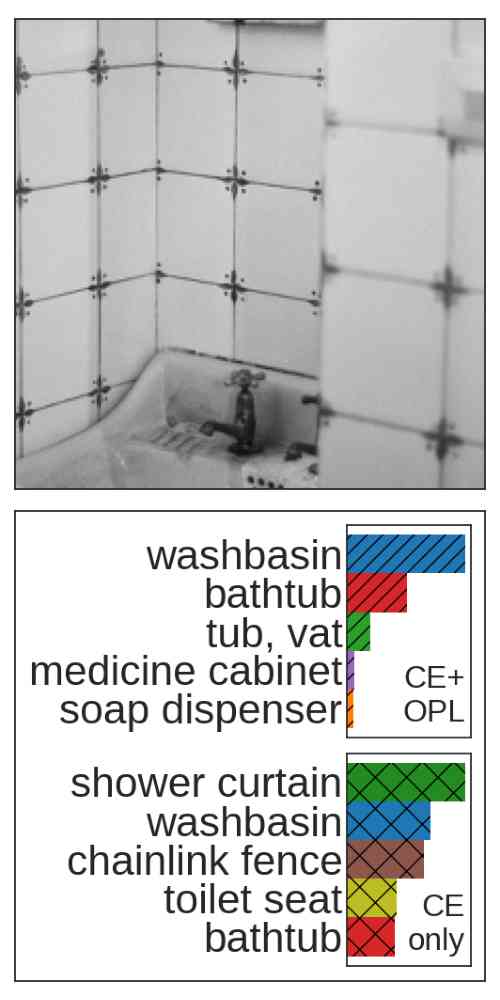}
    \includegraphics[width=0.24\linewidth]{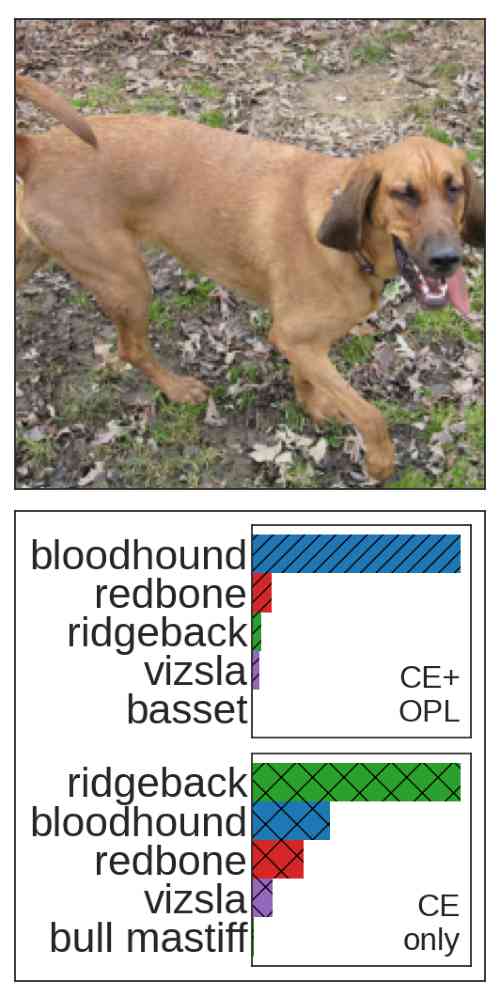}
    \includegraphics[width=0.24\linewidth]{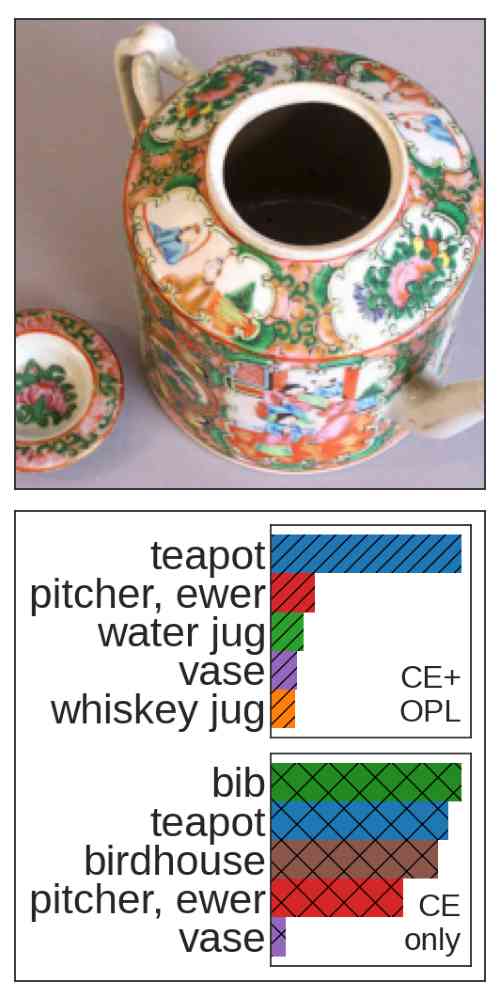}
    \includegraphics[width=0.24\linewidth]{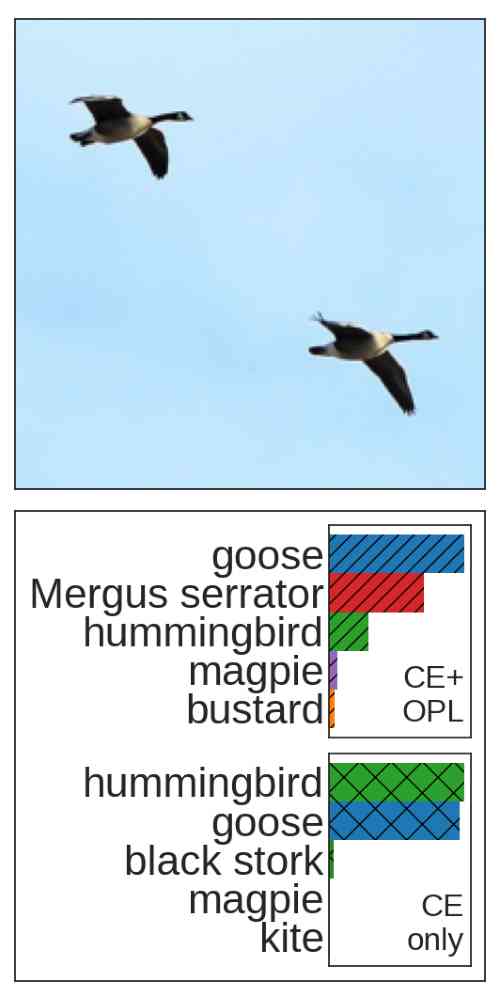}
    \includegraphics[width=0.24\linewidth]{figures/plots/images/joint_043.jpg}
    \includegraphics[width=0.24\linewidth]{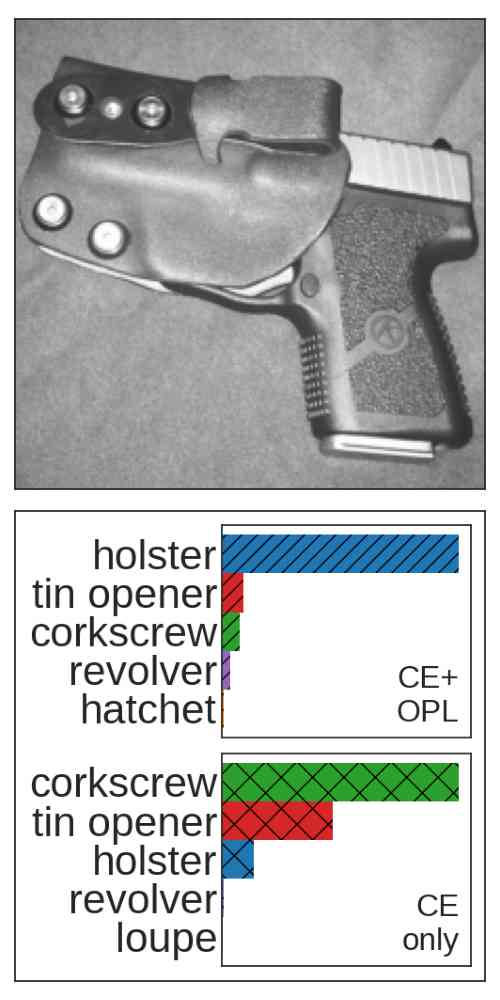}
    \includegraphics[width=0.24\linewidth]{figures/plots/images/joint_077.jpg}
    \caption{\textbf{Visualization of Classification Results:} We show some examples of images where OPL predicts the correct class but CE fails.}
    \label{fig:app_imnet_01}
\end{figure*}

We further explore the performance of our model (CE+OPL) trained on ImageNet (model used for experiments presented in Table~\ref{table:baseline_imagenet} of main paper) by examining the failure cases of the baseline model that were improved upon when adding OPL. Visualizations for some randomly selected such cases are illustrated in Fig.~\ref{fig:app_imnet_01} and Fig.~\ref{fig:app_imnet_02}. 

\begin{figure*}
    \centering
    \includegraphics[width=\linewidth]{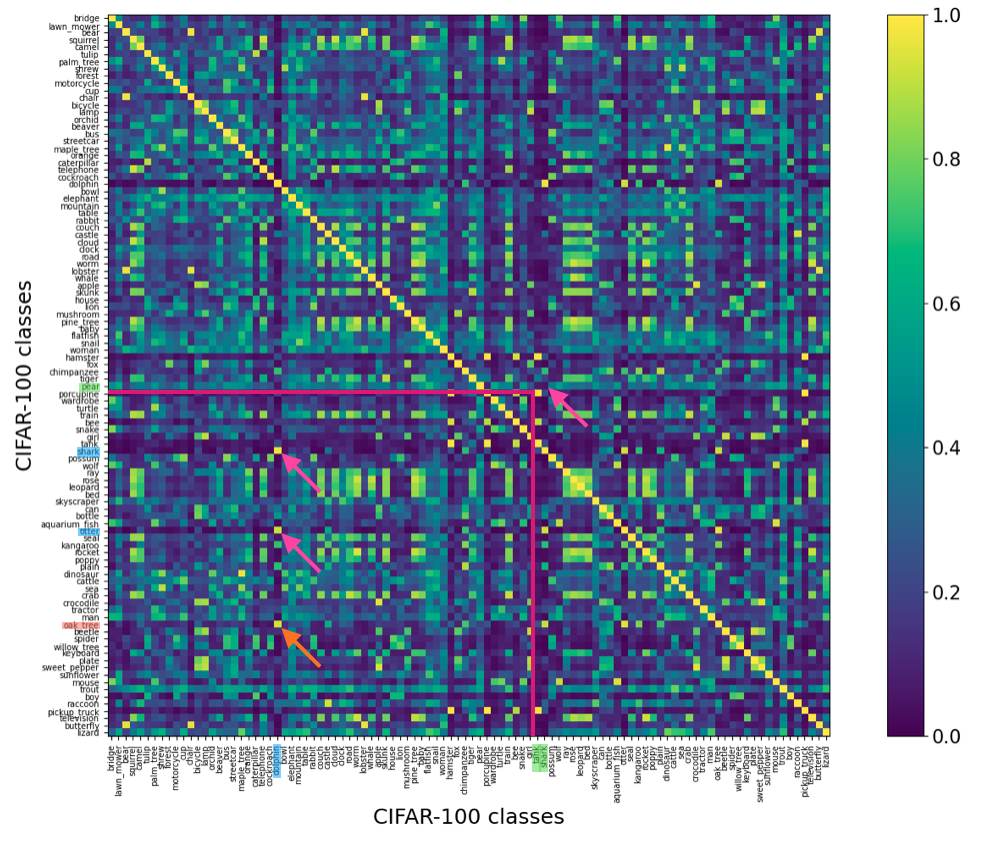}
	\caption{\textbf{Orthogonality Visualization:} The diagram (enlarged and elaborated version of Fig \ref{fig:ablation_block_plots_B} in main paper) visualizes the cosine similarity between each pair of per-class feature vectors extracted from an OPL trained ResNet-56 for the CIFAR-100 test-set. Each per-class feature vector is calculated averaging over the features of all samples belonging to that class within the test-set. We analyse the relationships for two randomly selected classes, \textit{dolphin} and \textit{pear}. Consider the similarity of the dolphin class column (label highlighted in blue). In general, it has low similarity with the other classes, except in 3 instances. Two of those, \textit{shark} and \textit{otter} (pink arrows) align with our heuristics on similarity of those categories. The similarity to \textit{oak tree} category can be attributed to some correlation present within the test-set images of these two classes (\eg, both contain large blue portions - ocean for \textit{dolphin} and sky for \textit{oak tree}). Now, consider pear (label highlighted in green), which has an average similarity to most other classes except two: \textit{tank} and \textit{shark} (labels highlighted in green / \textit{tank} in CIFAR-100 is the military vehicle). These two classes have relatively lower similarity with the \textit{pear} class as seen from the diagram (pink lines and pink arrow) which again aligns with our intuition about the relationships between these categories. Overall, we note that the orthogonality constraints we enforce on feature space through OPL allows room for learning hidden inter-class relationships which can be interpreted meaningfully, in comparison to the same relationships for the CE baseline.}
	\label{fig:block_plots_explained}
\end{figure*}

\subsection{Block Matrix}
\label{sec: block_matrix_analysis}
We defined the overall objective of OPL as a minimization of the expected inter-class orthogonality (refer Eq.~\ref{eq:opl_em} of main paper) and conducted empirical analysis using models training using our proposed loss function against a CE only baseline (illustrated in Fig.~\ref{fig:ablation_block_plots} of main paper). In this section, we conduct additional analysis on those block-matrices to further understand the outcomes of our orthogonality constraints on the learned feature space. It is interesting to note that while OPL enforces a higher degree of orthogonality between the average class vectors, it does not naively push everything to be orthogonal. We note that this allows any hidden knowledge learned during the training process (information not captured in the labels explicitly) to remain captured within the features. The results of the experiments conducted on this are illustrated in Fig.~\ref{fig:block_plots_explained}.

\end{document}